\documentclass{article}

\usepackage{bosonai_techreport}

\usepackage[utf8]{inputenc}
\usepackage[T1]{fontenc}
\usepackage[breaklinks=true,colorlinks=true,pageanchor=false]{hyperref}
\hypersetup{
  linkcolor={blue!50!black},
  citecolor={blue!50!black},
  urlcolor={blue!50!black},
}
\usepackage{url}
\usepackage{booktabs}
\usepackage{amsmath}
\usepackage{mdwlist}
\usepackage{amsfonts}
\usepackage{nicefrac}
\usepackage{microtype}
\usepackage{xcolor}
\usepackage{graphicx}
\usepackage{multirow}
\usepackage{subcaption}
\usepackage{enumitem}
\usepackage{tcolorbox}
\usepackage{tikz}
\usetikzlibrary{arrows.meta, positioning, fit, calc, decorations.pathreplacing}
\usepackage{pifont}
\tcbuselibrary{skins, breakable}

\newtcolorbox{promptbox}[1][]{
  colback=gray!5,
  colframe=gray!50,
  fonttitle=\bfseries\small,
  title={#1},
  breakable,
  enhanced,
  left=6pt, right=6pt, top=4pt, bottom=4pt,
  fontupper=\small\ttfamily,
}
\newcommand{\pvar}[1]{\textcolor{teal}{\textlangle #1\textrangle}}

\newtcolorbox{dialoguebox}[1][]{
  colback=blue!2,
  colframe=blue!25,
  fonttitle=\bfseries\small,
  title={#1},
  breakable,
  enhanced,
  left=6pt, right=6pt, top=4pt, bottom=4pt,
  fontupper=\small,
}
\newcommand{\passmark}{\textcolor{green!50!black}{\textbf{Pass}}}
\newcommand{\failmark}{\textcolor{red!70!black}{\textbf{Fail}}}
\newcommand{\partialmark}{\textcolor{orange!80!black}{\textbf{Partial}}}

\makeatletter
\renewcommand{\paragraph}{%
  \@startsection{paragraph}{4}{\z@}%
                {0.4ex \@plus 0.2ex \@minus 0.1ex}%
                {-0.8em}%
                {\normalsize\bfseries}%
}
\makeatother

\title{ProactBench: Beyond What The User Asked For}

\author{%
  Sepehr Harfi \hspace{1cm} Ahmad Salimi \\
  Boson AI \\
  Toronto, ON, Canada \\
  \texttt{\{sepehr, ahmad\}@boson.ai} \\
  \And
  Dongming Shen \hspace{1cm} Alex Smola\\
  Boson AI \\
  Santa Clara, CA, USA \\
  \texttt{\{dongming, smola\}@boson.ai} \\
}

\renewcommand{\bosonlogo}{boson}

\codelink{https://github.com/boson-ai/proactbench}
\datasetlink{https://huggingface.co/datasets/bosonai/proactbench}
\reportdate{May 2, 2026}

\begin{document}

\maketitle

\begin{abstract}
Most LLM benchmarks score how well a model responds to explicit requests. They leave unmeasured a different conversational ability: noticing and acting on needs the user has implied but not said. We call this \emph{conversational proactivity}. ProactBench decomposes it into three phase-tied types: \textsc{Emergent}, inference from a single disclosed anchor; \textsc{Critical}, synthesis across multiple anchors; and \textsc{Recovery}, grounded forward-looking value after task completion.

We operationalise the benchmark with three agents: a Planner, a User Agent, and an Assistant Model. Their information asymmetries defend against style-confounded scoring, rubric leakage, external-context contamination, and information dumps. The released corpus contains 198 curated dialogues with 624 trigger points across 24 communication styles drawn from a psychometric inventory and audited by an independent LLM judge. Across 16 frontier and open-weight models, \textsc{Recovery} is both difficult and weakly predicted by six standard benchmarks, making it a useful new evaluation signal.
\end{abstract}

\section{Introduction}
\label{sec:intro}

Benchmarks for LLMs mostly share the same setup: the user asks, and the model answers. MMLU \citep{hendrycks2021mmlu}, GPQA \citep{rein2024gpqa}, SWE-bench \citep{jimenez2024swebench}, LiveCodeBench \citep{jain2024livecodebench}, IFEval \citep{zhou2023ifeval}, and AIME \citep{aime2025} all fit this reactive form. Multi-turn evaluations such as MT-Bench \citep{zheng2023judging}, WildBench \citep{lin2024wildbench}, and Chatbot Arena \citep{chiang2024chatbot} preserve it: each turn is scored against the user's explicit ask.

As models improve on explicit requests, an important source of user value remains mostly unmeasured: whether the model notices and acts on needs the user has only implied. In real conversations, users rarely state every constraint, downstream use, risk, or next step. Some of the most useful assistance comes from identifying those unstated implications without being asked.

Consider a dialogue in which the user has spent several turns preparing for an event installation and signs off: \emph{``Packet's final. Sending tonight, then loading the hatchback right after work tomorrow, rush hour chaos.''} The task is complete. One model responds warmly: \emph{``Sounds like a clean plan. Drive safe, and go get it.''} A second model picks out a detail the user disclosed, loading the hatchback, and adds: \emph{``Pack the car in reverse order of installation. Your projector stand and extension cord go in last, so they're the first things out.''} Both responses are fluent; only the second turns a disclosed detail into useful forward-looking help. We call this difference \emph{conversational proactivity}.

Proactivity is not the same as doing the requested task well. \emph{Instruction-following} is reactive compliance, however polished, when the user's message already makes the action obvious. \emph{Clarification-seeking} is reactive when it asks about information the user has already supplied. \emph{Generic helpfulness}, including closing boilerplate such as ``let me know if you need anything,'' adds no grounded value. Proactive behaviour instead requires the model to act on an unstated implication, or to use a specific disclosed detail to introduce something useful that the user did not articulate.

This capability changes across a conversation. We treat these phase-tied behaviours as three trigger types: \textsc{Emergent}, inferring an unstated need from a single disclosed detail early on; \textsc{Critical}, synthesising several details into a new conclusion mid-dialogue; and \textsc{Recovery}, offering grounded forward-looking value after the user signals task completion. \textsc{Recovery} is especially important because task closure is where a model can sound helpful while saying nothing useful.

Proactive contributions of this kind are not a designer's preference. In a controlled pairwise study (Appendix~\ref{app:rubric_pairwise}), human raters prefer responses generated under our \textsc{Recovery} rubric to vanilla responses from the same model in $80\%$ of $144$ non-tie comparisons (95\% CI $[74\%, 86\%]$, $p < 10^{-12}$). What ProactBench scores is something users want to receive, not nanny-style overreach.

We introduce ProactBench, a benchmark for measuring this capability, and release data, code, and a comparison of 16 frontier and open-weight models. Our main contributions are:
\begin{enumerate}[leftmargin=*, itemsep=-2pt, topsep=0pt]
  \item We define conversational proactivity as a phase-tied capability with three trigger types: \textsc{Emergent}, \textsc{Critical}, and \textsc{Recovery}. The definition explicitly excludes instruction-following, clarification-seeking, generic helpfulness, and sycophantic closings, so the benchmark does not collapse into reactive competence (Section~\ref{sec:proactbench}).
  \item We introduce a three-agent evaluation architecture: a Planner, a User Agent, an Assistant Model, plus an offline judge. Their information asymmetries address four validity threats: style-confounded scoring, rubric leakage, external-context contamination, and information dumps (Sections~\ref{sec:design}--\ref{sec:validity}).
  \item We release a dataset of 198 curated dialogues with 624 triggers, grounded in Nemotron-Personas-USA \citep{meyer2025nemotronpersonas}, varied across 24 psychometric communication styles, and audited by an independent LLM judge (Section~\ref{sec:proactbench}).
  \item We evaluate 16 frontier and open-weight models and show that \textsc{Recovery} is weakly predicted by standard capability benchmarks. Long-context recall and sycophancy do not explain the gap; the remaining spread appears to reflect a policy difference, not only a capability difference (Section~\ref{sec:results}).
  \item In a pairwise human study, exposing the \textsc{Recovery} rubric to a model at generation time produces responses that human raters prefer to the same model's vanilla output in $80\%$ of $144$ non-tie comparisons. The proactive behaviour ProactBench defines and measures is something humans independently prefer (Section~\ref{sec:human_eval_results}, Appendix~\ref{app:rubric_pairwise}).
\end{enumerate}

\section{Related Work}
\label{sec:related}

\begin{description}[leftmargin=10pt, itemsep=0pt, topsep=0pt]
\item[Reactive and proactive benchmarks.]
Most widely used LLM evaluations score a response to an explicit request \citep{hendrycks2021mmlu, rein2024gpqa, jimenez2024swebench, jain2024livecodebench, zhou2023ifeval, aime2025}; multi-turn variants such as MT-Bench, WildBench, and Chatbot Arena preserve the same premise turn by turn \citep{zheng2023judging, lin2024wildbench, chiang2024chatbot}. A separate line studies proactivity, mostly in agentic settings: ProAgentBench \citep{tang2026proagentbench} and PARE \citep{nathani2026pare} score computer-use interventions, PROBE \citep{pasternak2025beyond} probes datastores for bottlenecks, and ProCIS \citep{samarinas2024procis} targets document retrieval. The closest dialogue benchmarks, ProactiveEval \citep{liu2025proactiveeval}, PROPER \citep{kaur2026proper}, and ProTOD \citep{dong2025protod}, score proactive moves but treat proactivity as a single scalar or as domain-specific gap filling. ProactBench instead separates when proactivity occurs in a conversation and varies user behaviour with a principled communication-style instrument.
\item[Adjacent capabilities.]
Conversational proactivity overlaps with several better-studied abilities but is not reducible to any of them. Theory-of-mind benchmarks \citep{le2019tomi, kim2023fantom, zhou2023sotopia} test mental-state attribution; \textsc{Emergent} and \textsc{Critical} require similar inference, but score whether it leads to a useful contribution in dialogue. Implicit-intent and active-elicitation work \citep{qian2024tellme, gan2024clarq, qian2025userbench, li2023gate, andukuri2024stargate} asks whether models uncover unstated needs, but usually stops once intent is resolved. \textsc{Recovery} asks what the model does after the user signals that the task is already complete. Sycophancy work \citep{sharma2023sycophancy, perez2022discovering, wei2023syntheticsycophancy} studies agreeable but epistemically weak behaviour; our \textsc{Recovery} rubric is designed to reject generic, agreeable closings. Long-context recall benchmarks \citep{liu2024lostmiddle, hsieh2024ruler, bai2024longbench} test retrieval, while our analysis shows that \textsc{Recovery} does not degrade with disclosure distance (Appendix~\ref{app:longctx}).
\item[Methodology.]
We build on LLM-as-judge evaluation \citep{zheng2023judging, lin2024wildbench, zhou2023sotopia} while addressing common failure modes: self-affinity \citep{panickssery2024selfrecognition}, inconsistency \citep{stureborg2024inconsistent}, positional bias \citep{wang2023notfair}, and evaluator quality \citep{bai2023examiner}. The key safeguards are prospective rubric declaration (Section~\ref{sec:design}) and a cross-family judge swap (Section~\ref{sec:judge_swap}). We also draw on persona-prompting work \citep{hu2024personaeffect, beck2024sociodemographic}, extending it to controlled behavioural variation across six CSI traits.
\end{description}

\section{ProactBench}
\label{sec:proactbench}

\subsection{Conversation design}
\label{sec:design}

A proactivity benchmark has to do more than score a static answer. It must create the conversation, decide when an unstated need has become testable, and score the response without letting the assistant model see the test. One corpus trigger illustrates the setup; the formal definition follows.

\begin{dialoguebox}[Example: \textsc{Emergent} trigger at turn 1, professional persona, chatty + questioning style]
\noindent\textbf{User:} \emph{``Quick gut check, what usually goes into a 60\% design review package for a small streetscape? \ldots\ Our drainage markups and BIM clash notes are still floating because the coordinator's out this week, so I'm trying to lock the outline before I caffeinate myself into chaos.''}

\vspace{1pt}\noindent\textbf{Qwen3.5-397B} [\passmark]: Since you're missing drainage and BIM coordination (which is risky at 60\%), \ldots\ if the coordinator is out, mark the drawings ``PRELIMINARY COORDINATION, CONFLICTS TO BE RESOLVED AT 90\%''. Add a Known Issues log \ldots\ Protect the schedule: make sure the client knows that if those coordination notes change significantly, the 90\% timeline might need a buffer.

\vspace{1pt}\noindent\textbf{GPT-5.5} [\partialmark]: If BIM clash notes are still floating, put in a plain-language coordination log so the review package still shows the issues. \ldots\ Items pending coordinator confirmation.
\end{dialoguebox}

\begin{figure}[t]
  \centering
  \includegraphics[width=\linewidth]{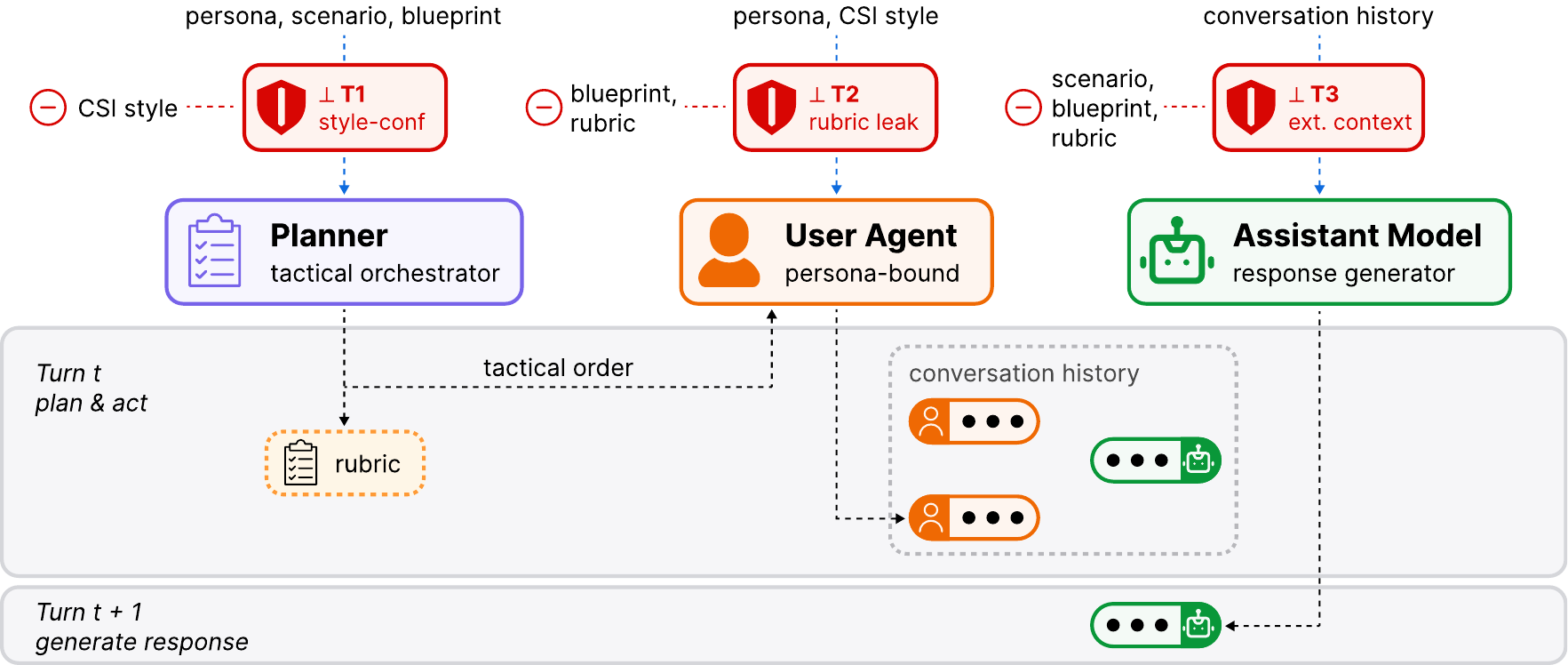}
  \caption{The three-agent loop. The Planner authors strategy and the prospective rubric for turn $t{+}1$. The User Agent renders the Planner's tactical orders as natural messages. The Assistant Model sees only the natural conversation.}
  \label{fig:architecture}
\end{figure}

The user discloses one factual detail, that the coordinator is absent, inside an otherwise closed request. We call such details \emph{anchors}. A \emph{trigger} is a turn whose response is scored against a rubric written before the model answers. The rubric specifies a pass criterion, a partial-credit criterion, and a fail criterion, all grounded in what the model has seen so far. In the example, the proactive principle is that an absent coordinator implies an unstated ownership and schedule risk. Qwen3.5-397B passes by surfacing the resourcing gap and schedule consequence; GPT-5.5 earns partial credit by noticing the coordination issue but mostly returning the requested checklist. Figure~\ref{fig:architecture} shows how the benchmark keeps this judgement prospective and hidden from the assistant model.
\begin{description}[leftmargin=10pt, itemsep=0pt, topsep=0pt]
\item[Planner.]
The Planner authors the dialogue strategy: which anchors to reveal, when to declare a trigger, and what rubric should govern that trigger. Trigger declarations are prospective. At turn $t$, before the assistant model answers, the Planner decides whether the response to that turn will be scored and commits the rubric. The Planner sees the persona, scenario, and dialogue history, but not the communication style.

\item[User Agent.]
The User Agent turns the Planner's tactical orders into natural user messages under an assigned persona and communication style.

\item[Assistant Model.]
The assistant model receives only the ordinary conversation: user messages and its own prior responses. It is not told that proactivity is being tested, and it never sees the trigger schedule, rubric, persona, or style.

\item[Offline judge.]
For model comparison, an offline judge reruns only the trigger turns of an existing dialogue. It sees the conversation history, the fixed rubric, and the regenerated response, but no persona or style. This lets every evaluated model face identical contexts and rubrics while avoiding another stochastic dialogue rollout.
\end{description} 

Formally, let $H_{1:t} = \{(u_1, a_1), \dots, (u_{t-1}, a_{t-1}), u_t\}$ denote the dialogue history at turn $t$. A response $a_t$ is proactive if it addresses a need implied, but not stated, by $H_{1:t}$. Anchors are the factual details in user turns $u_i$ that create those inferential opportunities. The assistant model receives only $H_{1:t}$: no persona, no rubric, no trigger marker, and no prior memory of the user.

\subsection{Validity}
\label{sec:validity}

The benchmark relies on multiple LLM agents, so validity depends on what each agent can and cannot see. Figure~\ref{fig:architecture} summarises the information flow. We design the loop around four threats.

\begin{description}[leftmargin=10pt, itemsep=0pt, topsep=2pt]
\item[T1 (style-confounded scoring)]
If scoring depends on the user's communication style, proactive content can be mistaken for persona-matching tone. The Planner does not receive the communication-style prompt, and the offline judge receives the rubric and conversation history without persona or style metadata. The Planner is therefore blind to style at dialogue time, so rubrics are written without knowledge of how the user sounds. However, the blueprint that the Planner receives was generated with the style as an input (Stage~2, Appendix~\ref{app:pipeline}), because disclosure pacing must adapt to terse vs.\ chatty registers---a terse user naturally reveals fewer details per turn. The style therefore shapes the conversational rhythm the Planner navigates, but not the rubric criteria, which are grounded solely in which anchors have been disclosed by the current turn.

\item[T2 (rubric leakage)]
If the assistant model sees the rubric, trigger schedule, or orchestration logic, it can game the benchmark. We therefore show it only the natural conversation history.

\item[T3 (external context)]
The assistant model generates the assistant side of the conversation during curation. If it receives the scenario, blueprint, or rubric, its responses would reflect knowledge of the hidden goal and anchor schedule, contaminating the conversation history that all 16 models later see during offline evaluation. We enforce cold-start isolation: the assistant model sees only the conversation history, with no access to the persona, scenario, blueprint, or rubric, so the curated dialogues contain only information that a real assistant would naturally have.

\item[T4 (information dumps)]
If all anchors are disclosed at once, proactivity collapses into multi-document question answering. The Planner enforces an anchor-drip rule of at most one primary anchor per user turn, so each trigger tests inference from a controlled information state.
\end{description}

\subsection{Scenario}
\label{sec:scenario}

A dialogue in ProactBench is defined by three ingredients: who the user is, what hidden goal structures the conversation, and how the user communicates.
\begin{description}[leftmargin=10pt, itemsep=0pt, topsep=2pt]
\item[Personas.]
We sample 50 profiles from Nemotron-Personas-USA \citep{meyer2025nemotronpersonas}, a synthetic dataset of US profiles grounded in real-world demographics. Each profile includes five life domains: Professional, Sports \& Fitness, Arts \& Culture, Travel \& Exploration, and Culinary. The Planner and User Agent receive the full persona. The assistant model receives none of it; it must infer the user's situation only from details disclosed in the dialogue.
\item[Scenarios and blueprints.]
For each persona-category pair, we generate a \emph{proactive scenario}: a hidden main goal, an explicit surface request, implicit anchors to reveal incrementally, and an ideal assistant trajectory. We then expand each scenario into a \emph{blueprint}: a turn-by-turn plan that controls anchor disclosure, trigger placement, and how the conversation adapts when the assistant is proactive or reactive. An independent cross-family LLM judge audits each blueprint for blank-slate integrity, logical necessity, persona alignment, and rubric clarity; failed blueprints are excluded.
\item[Communication styles.]
We derive user styles from six attributes of the Communication Styles Inventory \citep{devries2011csi}: expressiveness, preciseness, verbal aggressiveness, questioningness, emotionality, and impression manipulativeness. Of the $2^6 = 64$ binary combinations, we retain 24 psychologically coherent styles (filtering rationale and full list in Appendix~\ref{app:styles}). Each style specifies behavioural directives and length: 5--25 words for terse styles, 40--100 for chatty styles.
\end{description}

\begin{table}[t]
  \caption{Trigger types and rubric bars. Each dialogue contains 3--6 triggers across the three windows.}
  \label{tab:trigger_types}
  \centering
  \small
  \smallskip
  \begin{tabular}{@{}llp{8.5cm}l@{}}
    \toprule
    \textbf{Type} & \textbf{Window} & \textbf{Tests} & \textbf{Bar} \\
    \midrule
    \textsc{Emergent} & Turns 1--3 & Inference of an unstated need from a single disclosed anchor. & High \\
    \textsc{Critical} & Turns 4--7 & Synthesis across 2 or more disclosed anchors into a new conclusion. & High \\
    \textsc{Recovery} & Turns 8--10 & Grounded forward-looking value tied to a specific earlier detail, after the user signals task completion. & Moderate \\
    \bottomrule
  \end{tabular}
\end{table}

\subsection{Evaluation protocol}
\label{sec:eval_protocol}

\begin{description}[leftmargin=10pt, itemsep=0pt, topsep=2pt]
\item[Trigger types.]
Each dialogue contains 3--6 triggers distributed across three phase windows (Table~\ref{tab:trigger_types}). \textsc{Emergent} triggers (intended turns 1--3) test single-anchor inference; \textsc{Critical} (intended turns 4--7) tests multi-anchor synthesis; \textsc{Recovery} (intended turns 8--10) tests grounded forward-looking value after the user has signalled completion. The Planner is instructed to follow these windows; a small fraction of triggers drift one turn outside, and the released corpus carries each realised turn. Generic offers always fail.
\item[Scoring.]
Triggers are scored as Pass, Partial, or Fail. Pass requires the proactive principle to be met: single-anchor inference for \textsc{Emergent}, multi-anchor synthesis for \textsc{Critical}, and grounded forward value for \textsc{Recovery}. Partial credit captures narrower but substantive contributions. Fail covers reactive execution, generic closings, hallucinations, and ignored anchors. We report both pass rate and weighted score, with $\textrm{Pass}=1$, $\textrm{Partial}=0.5$, and $\textrm{Fail}=0$.
\item[Grounded rubrics and evidence.]
The Planner generates every rubric at trigger time, using only anchors disclosed so far. Rubrics state the proactive principle without requiring a particular output format. Every score must include a rationale and a verbatim evidence quote from the response, giving an auditable trail rather than an unfalsifiable holistic judgement.
\item[Scale.]
Each evaluated model is scored on 198 curated dialogues, yielding about 624 triggers per model: 201 \textsc{Emergent}, 232 \textsc{Critical}, and 191 \textsc{Recovery}. Dialogues run up to 10 turns and average 3.15 triggers.
\item[Two-phase evaluation.]
The curation phase, which generates the dialogues and trigger points, runs the full three-agent loop once with Gemini-2.5-Pro as the assistant model, producing fixed dialogue histories, trigger locations, and rubrics. The offline phase then evaluates new models by regenerating only at those trigger turns and scoring each response against the original rubric. This holds contexts and rubrics constant across models and avoids rerunning the stochastic Planner/User-Agent loop for every model.
\item[Caveats.]
Each model is sampled once at temperature $0.7$; rank stability under three independent runs of two representative models is verified in Appendix~\ref{app:pipeline}. The Planner and offline judge are both GPT-5.4, so we report a cross-family judge swap in Section~\ref{sec:judge_swap}. Offline scoring also reuses conversation histories generated with the curation model, which a more proactive model might have changed. A 20-blueprint re-curation under GPT-5.5 (Appendix~\ref{app:curation_contamination}) preserves cross-model rankings (Spearman~$\rho$ \citep{spearman1904correlation} $=0.80$ on \textsc{Overall}) and suggests that Gemini curation produces slightly harder contexts, making the reported scores conservative.
\end{description}

\subsection{Code and data}
\label{sec:release}

We release the orchestration pipeline, evaluation scripts, and corpus under the \textbf{Apache 2.0} licence; persona-derived content inherits the upstream Nemotron-Personas-USA CC-BY-4.0 licence. The release includes the 198-dialogue benchmark corpus, 624 trigger-point rubrics and labels, generated blueprints, validation decisions, Croissant 1.0 metadata \citep{akhtar2024croissant} with Responsible-AI fields, and a Datasheet for Datasets \citep{gebru2021datasheets}. An anonymized read-only mirror of the full repository is available at \url{https://anonymous.4open.science/r/ProactBench-81A3/README.md}; the public Hugging Face dataset URL will be disclosed in the camera-ready version. Appendices~\ref{app:datasheet} and \ref{app:licensing} give release and licensing details.

\section{Results}
\label{sec:results}

We evaluate 16 models under the offline protocol of Section~\ref{sec:eval_protocol}. The panel includes frontier closed-source models (GPT-5.5, Claude-Opus-4.7, Gemini-3.1-Pro / 2.5-Pro / 2.5-Flash, o4-mini, GPT-4o), frontier open-weight models (Qwen3.5-397B-A17B, Kimi-K2.6, DeepSeek-V4-Flash, Llama-4-Maverick, MiMo-V2.5-Pro), and smaller open-weight models (Qwen3.5-9B, Qwen2.5-7B-Instruct, Llama-3.2-8B-Instruct, Qwen3-1.7B). Identifiers, citations, and access dates are in Appendix~\ref{app:models}.

\subsection{Overall performance and decorrelation}
\label{sec:overall}
\label{sec:benchselect}

\begin{figure}[t]
  \centering
  \includegraphics[width=\linewidth]{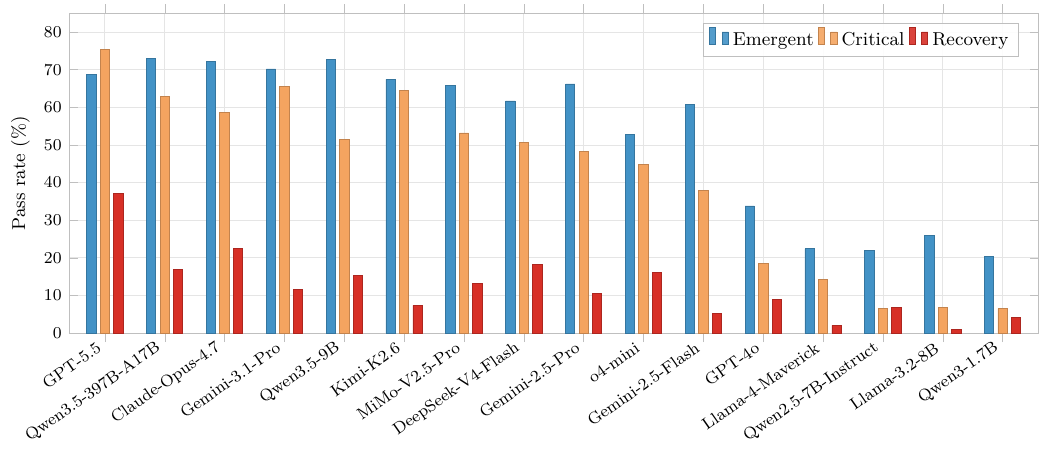}
  \caption{Per-model pass rate by trigger type. The drop from \textsc{Emergent} / \textsc{Critical} to \textsc{Recovery} is visible across all 16 models. Full per-type pass rates with bootstrap CIs are in Table~\ref{tab:main_results} in Appendix~\ref{app:models}.}
  \label{fig:trigger_bars}
\end{figure}

\textbf{Recovery is hard.} Figure~\ref{fig:trigger_bars} shows pass rates by trigger type across all 16 models. The best \textsc{Recovery} pass rate (GPT-5.5, $37.2\%$) is below the worst \textsc{Emergent} pass rate among frontier models (o4-mini, $52.7\%$). Llama-3.2-8B passes only $1.0\%$ of \textsc{Recovery} triggers, and 14 of 16 models pass fewer than $20\%$. Rankings also move sharply across types. Qwen3.5-397B leads \textsc{Emergent} ($72.9\%$) but drops to $17.0\%$ on \textsc{Recovery}, a 56-point gap. Kimi-K2.6 is competitive with GPT-5.5 on LiveCodeBench, SWE-bench Verified, and AIME 2025, yet scores $7.4\%$ on \textsc{Recovery}, five times worse than GPT-5.5. This suggests \textsc{Recovery} is not another proxy for benchmark strength.

\begin{figure}[t]
  \centering
  \begin{minipage}{0.59\linewidth}
    \includegraphics[width=\linewidth]{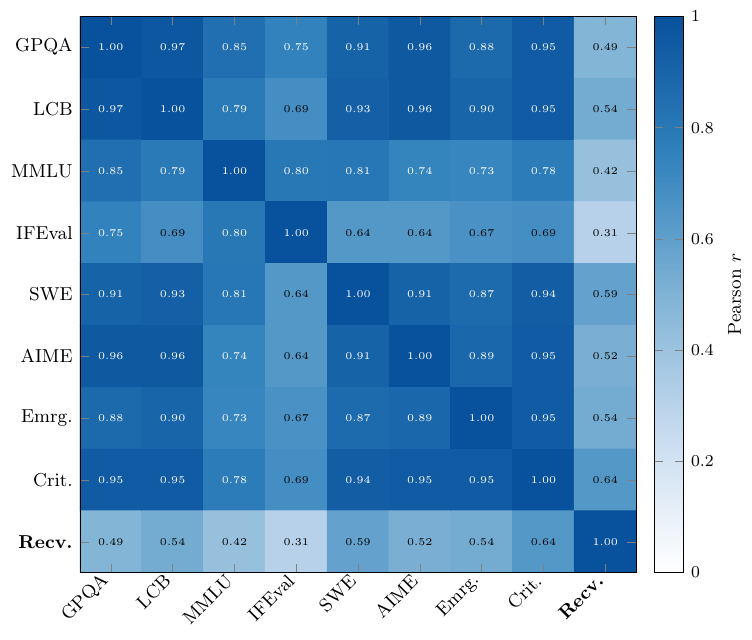}
  \end{minipage}\hfill
  \begin{minipage}{0.40\linewidth}
    \caption{Pairwise Pearson correlations across six standard benchmarks and three proactivity trigger types, computed over 16 models (95\% bootstrap CIs in Appendix~\ref{app:robustness_checks}). Existing benchmarks intercorrelate at $r = 0.64$ to $0.97$; \textsc{Emergent} and \textsc{Critical} fit within this regime ($\bar r \approx 0.83$). \textsc{Recovery} breaks the pattern: $\bar r = 0.51$, 95\% CI $[0.29, 0.71]$.}
    \label{fig:correlation}
  \end{minipage}
\end{figure}

\textbf{\textsc{Recovery} decorrelates from existing benchmarks.} The six standard benchmarks intercorrelate at $r = 0.64$ to $0.97$, reflecting a dominant general-capability factor (Figure~\ref{fig:correlation}). \textsc{Emergent} ($\bar r = 0.80$) and \textsc{Critical} ($\bar r = 0.86$) sit within that regime. \textsc{Recovery} does not: its mean correlation is $\bar r = 0.51$ (95\% CI $[0.29, 0.71]$). The bootstrap CI \citep{efron1979bootstrap} for the gap between the existing-benchmark mean ($0.83$) and the \textsc{Recovery} mean is $[0.08, 0.57]$, excluding zero. Under Holm--Bonferroni correction \citep{holm1979bonferroni} over the 36 pair tests, all six \textsc{Emergent} and \textsc{Critical} cross-benchmark correlations remain significant at $\alpha=0.05$; none of the six \textsc{Recovery} correlations do (Appendix~\ref{app:robustness_checks}). By submodular benchmark selection \citep{benchselect2026}, greedy entropy maximisation over the $9 \times 9$ correlation matrix ranks \textsc{Recovery} \#2 after GPQA Diamond and ahead of every other benchmark; over $10{,}000$ model-level resamples, \textsc{Recovery} is in the top three in $97\%$.

The contrast is clearest for Qwen3.5-397B-A17B and Kimi-K2.6. Compared with GPT-5.5 across six standard benchmarks (GPQA, LCB, MMLU, IFEval, SWE, and AIME), Kimi-K2.6 leads on LiveCodeBench ($+4.6$) and is competitive on SWE-bench Verified ($-1.8$) and AIME 2025 ($-3.6$), yet scores $5\times$ worse on \textsc{Recovery} ($7.4\%$ vs.\ $37.2\%$). No existing benchmark predicts this gap. Full per-model pass rates and weighted scores, with bootstrap CIs, are in Table~\ref{tab:main_results} of Appendix~\ref{app:models}.

\subsection{Per-stage performance: proprietary vs open-weight}
\label{sec:per_stage}

\textbf{Per-type spread.} Performance varies most sharply by trigger type. \textsc{Emergent} pass rate ranges from $20.4\%$ (Qwen3-1.7B) to $72.9\%$ (Qwen3.5-397B-A17B), roughly tracking general model strength. \textsc{Critical} ranges from $6.5\%$ to $75.4\%$ (GPT-5.5), with a steep drop below the 8B scale, consistent with multi-anchor synthesis requiring a capability threshold that smaller models often miss. \textsc{Recovery} ranges from $1.0\%$ (Llama-3.2-8B) to $37.2\%$ (GPT-5.5). Eight of 16 models move by three or more ranks between Overall and \textsc{Recovery}. Kimi-K2.6 is the clearest inversion: it ranks 6th Overall but 13th on \textsc{Recovery}, behind GPT-4o and Qwen2.5-7B despite outperforming both on every standard benchmark.

\textbf{Failure modes.} We categorise all 1{,}079 \textsc{Recovery} failures across eight representative models with a GPT-5.4 classifier. Two modes dominate: \emph{generic closing} ($44\%$) and \emph{deliverable repetition} ($45\%$); \emph{ungrounded suggestion} ($4\%$) and \emph{premature closure} ($6\%$) make up the tail. The two leaders split along a clear line: frontier closed-source models other than GPT-5.5 fail via generic closing (GPT-4o $70\%$, Qwen3.5-397B $61\%$, Gemini-2.5-Pro $61\%$), while GPT-5.5 itself and the smaller open-weight models fail via repetition (GPT-5.5 $69\%$, Qwen2.5-7B $83\%$, Qwen3-1.7B $83\%$). Appendix~\ref{app:failure_modes} gives the coding methodology and per-model breakdown.

\textbf{\textsc{Recovery} is not anti-sycophancy in disguise.} Among the eight failure-coded models, the within-failure rate of generic closing does not predict \textsc{Recovery} weighted score (Pearson $r = -0.23$, 95\% CI $[-0.88, 0.56]$); a partial correlation controlling for GPQA Diamond also straddles zero (Appendix~\ref{app:robustness_checks}). \textsc{Recovery} therefore does not reduce to a one-dimensional sycophancy axis. Nor is it retrieval-bottlenecked: per-trigger pass rate does not decline with disclosure distance for 15 of 16 models (Appendix~\ref{app:longctx}).

\subsection{Style Calibrated Performance}
\label{sec:logit}

\begin{figure}[t]
  \centering
  \includegraphics[width=\linewidth]{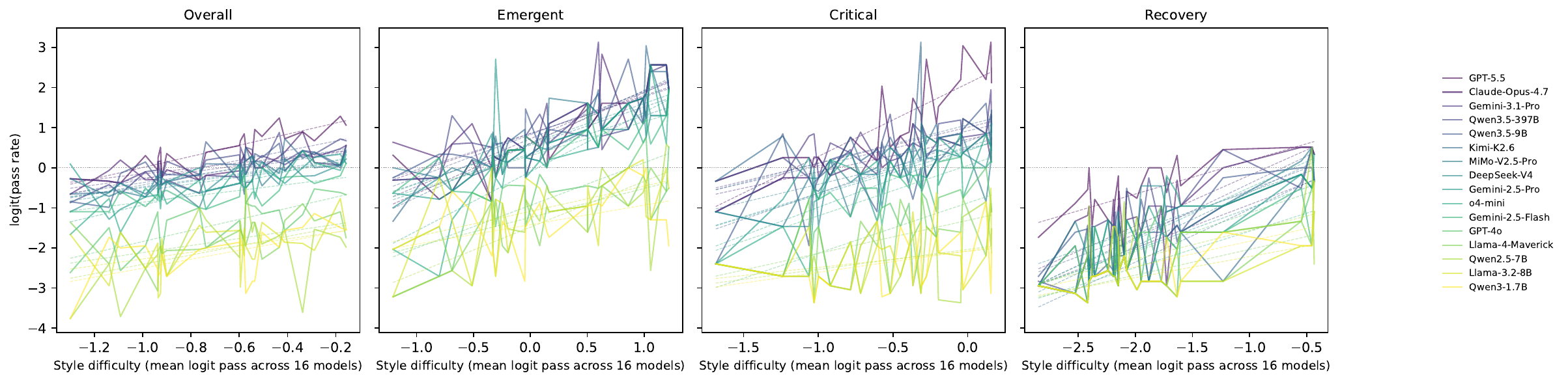}
  \caption{Logit-transformed per-(model, style) pass rates regressed against a shared style-difficulty axis (the cross-model mean logit per style), broken out by scoring axis. Each line is one of 16 models. They lie on approximately parallel lines for Overall, \textsc{Emergent}, and \textsc{Recovery} (slope std $\leq 0.27$); \textsc{Critical} is the widest-spread exception (slope std $0.36$). On \textsc{Recovery}, GPT-5.5's slope is $0.84$, below the cross-model mean, despite holding the highest absolute pass rate. The flat slope at the top of the leaderboard across easy and hard styles indicates a style-uniform policy.}
  \label{fig:logit}
\end{figure}

We next ask whether \textsc{Recovery} is a genuinely separate capability or simply a harder point on the same scaling curve. On raw pass rates, \textsc{Recovery} looks like an emergent ability \citep{wei2022emergent}: sub-8B models are near zero, while frontier models reach non-trivial rates. Following the caution of \citet{schaeffer2023mirage}, we test whether this visual threshold remains after using a continuous metric.

We compute a Laplace-smoothed logit for each model-style pass rate and regress it against a shared style-difficulty axis: the cross-model mean logit for that style. Under an item-response-style interpretation \citep{polo2024tinybenchmarks}, $\textrm{logit}(p_{m,c}) = a_m + b_m \,\bar\ell_c$, where $a_m$ captures overall ability and $\bar\ell_c$ captures style difficulty. If models differ only in ability, their fitted lines should be roughly parallel. The exact transform, a complementary per-stage check, and per-model coefficients are in Appendix~\ref{app:logit}.

For Overall, \textsc{Emergent}, and \textsc{Recovery}, the 16 models do lie on approximately parallel lines (slope std $\leq 0.27$). \textsc{Critical} is the wider-spread exception (slope std $0.36$). This means that, within a trigger type, user style mostly shifts difficulty in a common way across models.

The more interesting result is GPT-5.5's \textsc{Recovery} slope. It is $0.84$, below the cross-model mean, despite GPT-5.5 having the highest absolute \textsc{Recovery} pass rate. Other strong \textsc{Recovery} models (Qwen3.5-397B, Kimi-K2.6, MiMo-V2.5-Pro, Claude-Opus-4.7) have slopes above $1.2$. GPT-5.5's lead is therefore more style-uniform: it holds across easy and hard user styles. Its strongest competitors concentrate their advantage on easier styles and lose it under more hostile user behaviour. This is evidence that the \textsc{Recovery} gap reflects a policy difference, not just raw ability; we conjecture that post-training data mix and user-session feedback are plausible drivers.

\subsection{Human annotation}
\label{sec:human_eval_results}

A benchmark for proactivity is useful only if its labels track human judgement. We therefore ask human annotators to perform the same core task as the LLM judge: read the relevant dialogue context, apply the rubric, and assign Pass, Partial, or Fail.

We sample 60 trigger points stratified jointly by type (20 \textsc{Emergent} / 20 \textsc{Critical} / 20 \textsc{Recovery}), judge score within type (7 Pass / 7 Partial / 6 Fail), and evaluated model (12 items each across Claude-Opus-4.7, GPT-5.5, Gemini-3.1-Pro, Qwen3.5-397B-A17B, and Qwen3.5-9B). Sixteen Prolific slots were opened; 16 workers completed all 17 items in their slate, two contributed partial slates, and one was excluded under the pre-registered quality criterion ($\kappa_{\text{quad}} < 0.10$ vs.\ judge with $n \geq 5$). The remaining 17 raters contributed 258 ratings. Of the 60 sampled items, 45 received at least three ratings and entered the majority-vote consensus statistic. All bootstrap CIs use 10{,}000 item-level resamples (seed $2026$).

Krippendorff's $\alpha$ \citep{krippendorff2004reliability} (ordinal, items with $\geq 2$ raters) is $0.69\,${\tiny[$0.60$, $0.76$]}. By trigger type, $\alpha$ is $0.60\,${\tiny[$0.48$, $0.69$]} for \textsc{Emergent}, $0.61\,${\tiny[$0.49$, $0.71$]} for \textsc{Critical}, and $0.69\,${\tiny[$0.56$, $0.79$]} for \textsc{Recovery}. This indicates substantial reliability, with the strongest agreement on the post-completion behaviour that motivates the benchmark.

On the 45 consensus items, exact agreement with the GPT-5.4 judge is $29/45$ ($64.4\%$), within-one-step is $42/45$ ($93.3\%$), and only three items show a Pass$\leftrightarrow$Fail flip. Cohen's $\kappa_{\text{quad}}$ \citep{cohen1960kappa, cohen1968weighted} is $0.59\,${\tiny[$0.34$, $0.79$]} at the consensus level and $0.50\,${\tiny[$0.40$, $0.60$]} over all 258 ratings, with per-trigger consensus $\kappa_{\text{quad}}$ from $0.40$ (\textsc{Emergent}) to $0.79$ (\textsc{Recovery}). Disagreement is balanced (5 harsher / 11 lenient at consensus), indicating no systematic pass/fail bias.

A separate question is whether what we define as proactivity is actually valued by humans. A second, paired human study tests this directly: we expose the per-item rubric to gemini-2.5-pro as a system instruction and let it regenerate each \textsc{Recovery} response. Human raters prefer the rubric-conditioned response over vanilla generation in $80\%$ of 144 paired comparisons ($p < 10^{-12}$; Appendix~\ref{app:rubric_pairwise}), with preference rates above $70\%$ in every stratum including items where the vanilla response had already passed. The proactive behaviour our rubric targets is not just measurable---it is preferred. Full protocol, per-trigger confusion matrices, and the rubric-conditioning study details are in Appendices~\ref{app:human_eval}--\ref{app:rubric_pairwise}.

\subsection{Judge swap and model affinity}
\label{sec:judge_swap}

Because the Planner and offline judge are both GPT-5.4, the original ranking could reflect stylistic self-affinity rather than a true capability gap, especially for \textsc{Recovery}, where the OpenAI-family lead is largest. We bound this risk by rerunning offline evaluation on a 50-dialogue stratified subsample under two cross-family judges, Claude-Opus-4.7 and Kimi-K2.6. Holding all other inputs fixed, we re-judge seven models spanning capability and vendor diversity: GPT-5.5, Claude-Opus-4.7, Gemini-2.5-Pro, Qwen3.5-397B-A17B, Kimi-K2.6, DeepSeek-V4-Flash, and Qwen3.5-9B.

GPT-5.5 remains top-ranked on \textsc{Recovery} weighted score under all three judges, and \textsc{Recovery} remains the hardest trigger type for every model under every judge. Pairwise overall Cohen's $\kappa$ falls in the moderate range common for LLM-as-judge studies on subjective rubrics ($0.35$--$0.46$, all 95\% CIs strictly positive). Spearman $\rho$ on overall ranking is $0.64$--$0.89$ across the three judge pairs, reaching $\rho = 0.96$ on \textsc{Recovery}-specific rankings for the GPT-5.4 / Claude pair.

\textbf{Recovery gap compresses.} GPT-5.5's \textsc{Recovery} lead over the second-best model is $+25.5$\,pp under the original GPT-5.4 judge, $+8.2$\,pp under Claude-Opus-4.7, and $+5.0$\,pp under Kimi-K2.6. We treat the cross-family margins as more conservative. The qualitative claim is unchanged: GPT-5.5 leads on \textsc{Recovery}. Full protocol, $\kappa$ matrices, and per-cell tables are in Appendix~\ref{app:judge_swap_appx}.

\section{Discussion and Summary}
\label{sec:discussion}

\paragraph{Summary.}
Standard benchmarks ask whether a model answered the user's question; ProactBench asks whether it noticed what the user did not say. We separate this into three phase-tied trigger types and build an architecture (Planner writes the rubric before the model speaks; User Agent renders tactical orders; Assistant Model sees only the conversation) that defends against four named validity threats. The released benchmark, 198 dialogues with 624 triggers across 24 communication styles, ships under Apache-2.0 with persona-derived content under CC-BY-4.0. Sixteen frontier and open-weight models pass \textsc{Emergent} and \textsc{Critical} at rates close to existing capability benchmarks, but \textsc{Recovery} behaves differently: it decorrelates from six standard reasoning and coding benchmarks, no model passes more than $37\%$ of triggers, and the spread across models is not predicted by capability alone. Cross-family judge swaps and a curation-base swap preserve the rankings; a Prolific human-validation study finds substantial agreement with the GPT-5.4 judge (Krippendorff's $\alpha = 0.69$, consensus $\kappa_{\text{quad}} = 0.59$); a second Prolific study confirms that the proactive behaviour ProactBench targets is something humans prefer, with rubric-conditioned responses chosen over vanilla in $80\%$ of $144$ non-tie comparisons. The \textsc{Recovery} deficit is a behavioural default, not a capability ceiling.

\paragraph{Future work.}
Per-trigger rubrics, evidence quotes, and Pass/Partial/Fail labels make the corpus a substrate for reward modelling and preference-pair construction in the post-completion regime, where supervised data is sparse. The rubric-conditioning result (Appendix~\ref{app:rubric_pairwise}) gives a concrete recipe: vanilla and rubric-conditioned generations from the same model form pre-labelled preference pairs at scale, with the human-preferred direction known by construction. The goal is not to train models to be maximally proactive but to learn when initiative is welcome; pairing ProactBench with real user-session preference data that captures that distinction would make it actionable. A richer user simulator with longer horizons, multi-session memory, and varied disengagement patterns is a second direction.

\paragraph{Weak points.}
Three caveats matter. First, personas are US-English from a single source \citep{meyer2025nemotronpersonas}; norms around unsolicited advice differ across languages, cultures, and deployment contexts, so ProactBench is a capability probe, not a universal training target. Second, the dialogues, personas, and styles are synthetic, which gives us control but not real-user variability. Third, scoring is by an LLM judge with ranking-level reliability and only moderate per-trigger agreement (at least relative to large open-weights models); human ratings and real session data would be helpful to calibrate the judge further.

\newpage
\bibliographystyle{plainnat}
\bibliography{references}


\newpage
\appendix

\section*{Appendix}

The appendix is intended to make the benchmark easy to audit and reuse. It begins with the ingredients of the evaluation--the models, personas, communication styles, and generation pipeline--then gives worked examples, validation studies, robustness checks, and release details. The list below is a reader's map of where each kind of evidence lives.

\begin{description}[leftmargin=1.5em, itemsep=1pt, topsep=2pt]
  \item[\ref{app:models}\;\;Models.] The 16 evaluated models, their access routes, citations, and the full per-model results table with bootstrap CIs.
  \item[\ref{app:personas}\;\;Personas.] How personas are sampled from Nemotron-Personas-USA, what fields are used, how categories are represented, and what a scenario record looks like.
  \item[\ref{app:styles}\;\;Communication style.] The six CSI dimensions, the 24 retained binary combinations, the rationale for dropping the other 40, and per-style trigger counts.
  \item[\ref{app:pipeline}\;\;Pipeline.] The end-to-end generation and scoring pipeline, including blueprint auditing, persona / scenario / blueprint / dialogue counts, repeat-run rank stability, and token usage.
  \item[\ref{app:examples}\;\;Examples.] Planner and User Agent system-prompt excerpts, followed by three full example dialogues with rubrics and per-turn scores.
  \item[\ref{app:human_eval}\;\;Human calibration.] The Prolific validation protocol ($n = 60$ stratified trigger points, 16 completed slates, 258 ratings retained), including recruitment, interface, quality criteria, analysis plan, and full per-trigger agreement results.
  \item[\ref{app:judge_swap_appx}\;\;Judge swap.] Cross-family judge protocol, per-judge scores, pairwise Cohen's $\kappa$ and Spearman $\rho$ matrices, the \textsc{Recovery} gap analysis, the per-cell heatmap, and the curation-model contamination ablation.
  \item[\ref{app:failure_modes}\;\;Failure modes.] The coding scheme for four \textsc{Recovery} failure modes, per-model breakdowns for the eight failure-coded models, and discussion of frontier-vs-small failure profiles.
  \item[\ref{app:rubric_pairwise}\;\;Rubric-conditioned generation.] A pre-registered pairwise human study comparing vanilla and rubric-conditioned \textsc{Recovery} responses from the same model. Reports a B-preference rate of $0.80$ on $144$ non-tie comparisons, with per-stratum breakdowns and the pre-registered attentiveness filter.
  \item[\ref{app:novelty}\;\;Novelty of ProactBench.] A side-by-side comparison with adjacent benchmarks in proactivity, theory-of-mind, implicit intent, sycophancy, and long-context recall.
  \item[\ref{app:robustness_checks}\;\;Statistical robustness checks.] Multiple-comparison correction over the 36 cross-benchmark pair tests, bootstrap stability of the greedy-entropy ordering, and the sycophancy cross-correlation analysis.
  \item[\ref{app:longctx}\;\;Long-context-recall analysis.] A per-model regression of weighted \textsc{Recovery} pass rate on disclosure distance, with the full slope table.
  \item[\ref{app:logit}\;\;Logit transform.] The Laplace-smoothed logit transform, the IRT-style modelling rationale, the per-stage consistency check, and per-model fit coefficients.
  \item[\ref{app:datasheet}\;\;Datasheet.] The Gebru-style datasheet for the released corpus, plus pointers to the Croissant 1.0 metadata and Responsible-AI fields.
  \item[\ref{app:licensing}\;\;Licensing and \texttt{CITATION}.] Apache-2.0 release terms, persona-derived CC-BY-4.0 inheritance from Nemotron-Personas-USA, and the canonical \texttt{CITATION.cff}.
\end{description}

\section{Models}
\label{app:models}


The model appendix pins each reported score to the concrete system that produced it. We group the 16 evaluated models into three capability tiers, not to impose a strict ordering, but to make the comparison easier to read:
\begin{description}[leftmargin=10pt, itemsep=0pt, topsep=2pt]
\item[Frontier closed-source.] GPT-5.5 \citep{openai_gpt55_2026}, Claude-Opus-4.7 \citep{anthropic_opus47_2026}, Gemini-3.1-Pro, Gemini-2.5-Pro, and Gemini-2.5-Flash \citep{deepmind_gemini_2025}, o4-mini \citep{openai_o3_o4_mini_2025}, GPT-4o \citep{openai_gpt4o_2024}.
\item[Frontier open-weight.] Qwen3.5-397B-A17B \citep{qwen35_2026}, Kimi-K2.6 \citep{moonshot_kimik26_2026}, DeepSeek-V4-Flash \citep{deepseek_v4_2026}, Llama-4-Maverick \citep{meta_llama4_2025}, MiMo-V2.5-Pro \citep{mimo_2026}.
\item[Smaller open-weight.] Qwen3.5-9B \citep{qwen35_2026}, Qwen2.5-7B-Instruct \citep{qwen25_2024}, Llama-3.2-8B-Instruct \citep{meta_llama32_2024}, Qwen3-1.7B \citep{yang2025qwen3}.
\end{description}

\paragraph{Endpoints and access dates.}
Table~\ref{tab:model_versions} is the source of truth for model identifiers, access routes, and access dates. API-accessed models are queried through the vendor's default chat-completions interface; local models are served from HuggingFace checkpoints on a single 8$\times$A100-80GB GPU node via the Transformers $+$ Accelerate stack. Kimi-K2.6, Llama-4-Maverick, and MiMo-V2.5-Pro are accessed through OpenRouter via the OpenAI-compatible endpoint described in Section~\ref{sec:eval_protocol}.

\begin{table}[ht]
  \caption{Endpoints, access routes, and access dates for the 16 evaluated models. \emph{API} denotes the vendor's hosted chat-completions endpoint; \emph{local} denotes inference from a HuggingFace checkpoint on our hardware; \emph{OpenRouter} denotes a cross-vendor OpenAI-compatible aggregator. The curation phase used \texttt{gemini-2.5-pro} as the assistant model; all 16 listed below are scored offline against the same 198-dialogue corpus (Section~\ref{sec:eval_protocol}).}
  \label{tab:model_versions}
  \centering
  \smallskip
  \begin{tabular}{@{}llll@{}}
    \toprule
    \textbf{Model} & \textbf{Identifier} & \textbf{Access} & \textbf{Date} \\
    \midrule
    GPT-5.5            & \texttt{gpt-5.5}                            & API        & 2026-04 \\
    Claude-Opus-4.7    & \texttt{claude-opus-4-7}                    & API        & 2026-04 \\
    Gemini-3.1-Pro     & \texttt{gemini-3.1-pro}                     & API        & 2026-04 \\
    Gemini-2.5-Pro     & \texttt{gemini-2.5-pro}                     & API        & 2026-04 \\
    Gemini-2.5-Flash   & \texttt{gemini-2.5-flash}                   & API        & 2026-04 \\
    o4-mini            & \texttt{o4-mini-2025-04-16}                 & API        & 2026-04 \\
    GPT-4o             & \texttt{gpt-4o-2024-11-20}                  & API        & 2026-04 \\
    Qwen3.5-397B-A17B  & \texttt{qwen3.5-397b-a17b-instruct}         & API        & 2026-04 \\
    Kimi-K2.6          & \texttt{moonshotai/kimi-k2.6}               & OpenRouter & 2026-04 \\
    DeepSeek-V4-Flash  & \texttt{deepseek-ai/DeepSeek-V4-Flash}      & API        & 2026-04 \\
    Llama-4-Maverick   & \texttt{meta-llama/llama-4-maverick}        & OpenRouter & 2026-04 \\
    MiMo-V2.5-Pro      & \texttt{xiaomi/mimo-v2.5-pro}               & OpenRouter & 2026-04 \\
    Qwen3.5-9B         & \texttt{qwen3.5-9b-instruct}                & API        & 2026-04 \\
    Qwen2.5-7B-Instruct& \texttt{Qwen/Qwen2.5-7B-Instruct}           & local      & 2026-04 \\
    Llama-3.2-8B       & \texttt{meta-llama/Llama-3.2-8B-Instruct}   & local      & 2026-04 \\
    Qwen3-1.7B         & \texttt{Qwen/Qwen3-1.7B}                    & local      & 2026-04 \\
    \bottomrule
  \end{tabular}
\end{table}

\paragraph{Decoding configuration.}
We use a simple, shared decoding policy wherever the serving interface allows it. API-accessed standard chat models are called with temperature $0.7$ and the vendor's default system prompt. Local models are decoded with nucleus sampling ($p{=}0.9$, temperature $0.7$). Reasoning models (o4-mini and Gemini-2.5-Pro under thinking mode) require temperature $1.0$ and ignore $\text{top-}p$; for these we use \texttt{reasoning\_effort = medium} (or the vendor equivalent) and a 32k completion-token budget. Each model is sampled \emph{once} per trigger; the repeat-run check below measures how much this choice affects rankings.

\paragraph{Compute and runtime.}
The benchmark is small enough to rerun, but not trivial. A full 198-dialogue offline run takes roughly 40--90 minutes per local model on a single 8$\times$A100-80GB GPU node depending on parameter count, and 15--30 minutes per API model with 8 concurrent worker threads. Per-model token totals (assistant prompt $+$ completion, plus the corresponding judge call totals) are tabulated in Appendix~\ref{app:pipeline}.

\paragraph{Reproducibility.}
To check whether single-sample decoding changes the story, we rerun GPT-5.5 and Qwen3.5-9B three times each, choosing one frontier model and one mid-scale model. Per-trigger-type rankings are preserved across all three runs, and overall weighted scores vary by less than $1.6$ percentage points. The full log is in Appendix~\ref{app:pipeline}.

\paragraph{Per-model pass rates.}
Table~\ref{tab:main_results} gives the full numerical companion to Figure~\ref{fig:trigger_bars}: per-type pass rates and weighted scores for all 16 models, with 95\% bootstrap confidence intervals.

\begin{table}[t]
  \caption{Main results: pass rate (\%) and weighted score taking partial passes into acount (\%) by trigger type across 16 models, sorted by overall pass rate. Each model is evaluated on 198 dialogues ($\sim$201 \textsc{Emergent}, $\sim$232 \textsc{Critical}, $\sim$191 \textsc{Recovery} triggers). 95\% bootstrap confidence intervals (1\,000 trigger-level resamples) shown in brackets below each pass rate. \textsc{Recovery} is dramatically harder than the other types: even GPT-5.5 (the leading model) fails 63\% of the time, and 14 of 16 models pass fewer than 20\% of \textsc{Recovery} triggers.}
  \label{tab:main_results}
  \centering
  \small
  \setlength{\tabcolsep}{4pt}
  \begin{tabular}{@{}l cc cc cc cc@{}}
    \toprule
    & \multicolumn{2}{c}{\textbf{Overall}} & \multicolumn{2}{c}{\textbf{Emergent}} & \multicolumn{2}{c}{\textbf{Critical}} & \multicolumn{2}{c}{\textbf{Recovery}} \\
    \cmidrule(lr){2-3} \cmidrule(lr){4-5} \cmidrule(lr){6-7} \cmidrule(lr){8-9}
    \textbf{Model} & Pass [CI] & Wt. & Pass [CI] & Wt. & Pass [CI] & Wt. & Pass [CI] & Wt. \\
    \midrule
    GPT-5.5                & \textbf{61.5}\,{\tiny[57.4,65.4]} & \textbf{69.8} & 68.7\,{\tiny[62.7,75.1]} & 73.1 & \textbf{75.4}\,{\tiny[69.8,80.6]} & \textbf{83.2} & \textbf{37.2}\,{\tiny[30.4,44.0]} & \textbf{50.0} \\
    Qwen3.5-397B-A17B      & 52.5\,{\tiny[48.6,56.3]} & 60.6 & \textbf{72.9}\,{\tiny[66.3,78.9]} & 76.9 & 63.0\,{\tiny[57.0,68.7]} & 73.5 & 17.0\,{\tiny[12.1,22.5]} & 26.6 \\
    Claude-Opus-4.7        & 51.9\,{\tiny[48.1,55.8]} & 60.6 & 72.1\,{\tiny[66.2,78.1]} & \textbf{78.6} & 58.6\,{\tiny[52.6,64.7]} & 70.5 & 22.5\,{\tiny[16.8,28.8]} & 29.6 \\
    Gemini-3.1-Pro         & 50.4\,{\tiny[46.5,54.3]} & 58.3 & 70.2\,{\tiny[63.1,76.8]} & 77.3 & 65.5\,{\tiny[59.5,71.6]} & 76.5 & 11.5\,{\tiny[7.3,16.2]} & 16.5 \\
    Qwen3.5-9B             & 48.0\,{\tiny[43.6,52.2]} & 57.0 & 72.8\,{\tiny[66.1,78.9]} & 77.2 & 51.4\,{\tiny[44.3,58.4]} & 64.6 & 15.4\,{\tiny[10.3,21.2]} & 24.7 \\
    Kimi-K2.6              & 47.5\,{\tiny[43.6,51.6]} & 54.3 & 67.4\,{\tiny[60.6,73.6]} & 73.1 & 64.4\,{\tiny[58.2,70.7]} & 73.6 & 7.4\,{\tiny[3.7,11.6]} & 12.4 \\
    MiMo-V2.5-Pro          & 44.9\,{\tiny[40.9,48.8]} & 52.8 & 65.7\,{\tiny[59.7,72.1]} & 71.6 & 53.2\,{\tiny[47.2,59.3]} & 64.3 & 13.1\,{\tiny[8.4,17.8]} & 19.1 \\
    DeepSeek-V4-Flash      & 44.3\,{\tiny[40.6,48.2]} & 53.5 & 61.7\,{\tiny[55.7,68.2]} & 66.7 & 50.6\,{\tiny[44.2,57.1]} & 64.5 & 18.3\,{\tiny[13.6,24.1]} & 26.4 \\
    Gemini-2.5-Pro         & 42.5\,{\tiny[38.5,46.6]} & 50.3 & 66.2\,{\tiny[59.7,72.1]} & 73.1 & 48.3\,{\tiny[41.8,55.2]} & 62.1 & 10.5\,{\tiny[6.3,14.7]} & 12.0 \\
    o4-mini                & 38.6\,{\tiny[35.1,42.8]} & 47.4 & 52.7\,{\tiny[46.3,59.7]} & 57.7 & 44.8\,{\tiny[38.4,51.3]} & 58.8 & 16.2\,{\tiny[11.0,22.0]} & 22.5 \\
    Gemini-2.5-Flash       & 35.3\,{\tiny[31.4,38.9]} & 44.2 & 60.7\,{\tiny[53.2,67.2]} & 65.7 & 37.9\,{\tiny[31.5,44.0]} & 55.0 & 5.2\,{\tiny[2.6,8.4]} & 8.6 \\
    GPT-4o                 & 20.5\,{\tiny[17.3,23.6]} & 31.3 & 33.8\,{\tiny[27.9,40.3]} & 42.5 & 18.5\,{\tiny[13.8,23.7]} & 36.0 & 8.9\,{\tiny[5.2,12.6]} & 13.9 \\
    Llama-4-Maverick       & 13.1\,{\tiny[10.6,15.9]} & 22.5 & 22.4\,{\tiny[16.4,28.4]} & 29.1 & 14.2\,{\tiny[9.9,19.0]} & 31.9 & 2.1\,{\tiny[0.5,4.2]} & 4.2 \\
    Qwen2.5-7B-Instruct    & 11.5\,{\tiny[9.0,14.1]} & 21.6 & 21.9\,{\tiny[15.9,27.4]} & 31.8 & 6.5\,{\tiny[3.9,9.9]} & 17.9 & 6.8\,{\tiny[3.7,10.5]} & 15.4 \\
    Llama-3.2-8B           & 11.2\,{\tiny[8.7,13.6]} & 20.5 & 25.9\,{\tiny[19.9,32.8]} & 37.3 & 6.9\,{\tiny[3.9,10.3]} & 18.8 & 1.0\,{\tiny[0.0,2.6]} & 5.0 \\
    Qwen3-1.7B             & 10.3\,{\tiny[7.9,12.7]} & 22.0 & 20.4\,{\tiny[14.9,25.9]} & 31.8 & 6.5\,{\tiny[3.4,9.5]} & 21.1 & 4.2\,{\tiny[1.6,7.3]} & 12.8 \\
    \bottomrule
  \end{tabular}
\end{table}

\begin{table}[t]
  \caption{Frontier models that match or outperform GPT-5.5 on standard benchmarks but trail by 2--5$\times$ on \textsc{Recovery}. Qwen3.5-397B is competitive with GPT-5.5 on LiveCodeBench and SWE-bench; Kimi-K2.6 leads on LiveCodeBench, SWE-bench Verified, and AIME 2025, yet both have a \textsc{Recovery} pass rate below 20\%, while GPT-5.5 reaches 37\%. No existing benchmark would predict either inversion.}
  \label{tab:contrast}
  \centering
  \begin{tabular}{@{}lcccccc c@{}}
    \toprule
    & \multicolumn{6}{c}{\textbf{Standard benchmarks (\%)}} & \textbf{Ours} \\
    \cmidrule(lr){2-7} \cmidrule(lr){8-8}
    & GPQA & LCB & MMLU & IFEval & SWE & AIME & Recv. \\
    \midrule
    GPT-5.5 & \textbf{93.6} & 85.0 & \textbf{88.0} & \textbf{95.5} & \textbf{82.0} & \textbf{100.0} & \textbf{37.2} \\
    Qwen3.5-397B-A17B & 88.4 & 83.6 & 87.8 & 92.6 & 76.4 & 91.3 & 17.0 \\
    Kimi-K2.6 & 90.5 & \textbf{89.6} & 81.1 & 89.8 & 80.2 & 96.4 & 7.4 \\
    \bottomrule
  \end{tabular}
\end{table}

\section{Personas}
\label{app:personas}

\paragraph{Source and licence.}
ProactBench needs users with stable histories, preferences, and constraints, but it should not expose real people's profiles. We therefore seed the persona pool from \textbf{Nemotron-Personas-USA} \citep{meyer2025nemotronpersonas}, a synthetic dataset of US-based personality profiles released by NVIDIA. Each profile is generated from a probabilistic graphical model conditioned on real-world demographic and geographic distributions (US Census-derived age, sex, education, occupation, and location). This gives the benchmark a realistic aggregate population while keeping individual rows synthetic. Nemotron-Personas-USA is distributed under the \textbf{CC-BY-4.0} licence; ProactBench's persona-derived content inherits that licence. We chose a synthetic source rather than scraped real-user profiles to avoid privacy and re-identification risks at release time.

\paragraph{Per-profile fields used.}
Each Nemotron-Personas-USA row carries a short \emph{core personality} summary plus five life-domain aspects: \emph{professional / career}, \emph{sports \& fitness}, \emph{arts \& culture}, \emph{travel \& exploration}, and \emph{culinary \& food}. The Planner and User Agent receive the full multi-aspect persona at every turn (rendered by \texttt{\_build\_global\_persona} in \texttt{proactbench/prompts/synthesis.py}), so generated conversations can draw on a stable user background. The assistant model sees neither the persona nor the communication style. It must build its user model only from what the user discloses during the conversation. This is deliberate: in a cold-start deployment, that background would not normally be available unless the assistant had a separate CRM or personalization memory system, which is outside the scope of this paper and would confound the analysis.

\paragraph{Sampling.}
We stream the first \textbf{50 personas} from the dataset's \texttt{train} split, using deterministic order and no rejection sampling. Stage~1 generates candidate proactive scenarios across the five life-domain aspects per persona (\texttt{tasks.jsonl}). A curation pass then retains a stratified scenario base of \textbf{25 (persona, category) scenarios drawn from 19 personas}, exactly five per life-domain category, written to \texttt{selected\_tasks.jsonl}; remaining candidates are dropped at this step for redundancy with retained scenarios in the same category, cold-start infeasibility (reliance on real-time information the assistant cannot access), or persona-aspect imbalance, before any blueprint or audit cost is incurred. The 31 personas not represented in the retained set contributed candidate scenarios that were superseded by stronger candidates within the same category quotas. Stage~2 then renders each retained (persona, category) scenario under \textbf{10 communication-style combinations} (Appendix~\ref{app:styles}), yielding $25 \times 10 = 250$ blueprint candidates (Section~\ref{sec:scenario}). After independent-judge audit (210 PASS of 250) and Stage~4 rollout, $\mathbf{198}$ \textbf{dialogues} from these 19 personas remain in the released corpus. Table~\ref{tab:persona_categories} reports the released-corpus counts by category. The category imbalance is an outcome of the audit and rollout stages, not an additional sampling choice: Travel and Culinary scenarios fail more often because they are more likely to depend on real-time information the cold-start assistant cannot access.

\begin{table}[ht]
  \caption{Released-corpus dialogue counts per persona category. The 250 candidate (persona, category) pairs are reduced to 198 final dialogues by the independent-judge blueprint audit and a small number of curation-time failures (Section~\ref{sec:scenario}).}
  \label{tab:persona_categories}
  \centering
  \smallskip
  \begin{tabular}{@{}lr@{}}
    \toprule
    \textbf{Persona category} & \textbf{Dialogues} \\
    \midrule
    Professional / Career      & 46 \\
    Arts \& Culture            & 45 \\
    Culinary \& Food           & 37 \\
    Sports \& Fitness          & 37 \\
    Travel \& Exploration      & 33 \\
    \midrule
    \textbf{Total}             & \textbf{198} \\
    \bottomrule
  \end{tabular}
\end{table}

\paragraph{Scenario fields.}
For each (persona, category) pair, the Stage 1 generator (prompt: \texttt{PersonaProactivityScenarioPromptConfig} in \texttt{proactbench/prompts/synthesis.py}) produces a structured \texttt{ProactiveScenario}. The fields below separate what the user ultimately wants, what the assistant model is initially asked to do, and what evidence may surface later in the conversation:
\begin{description}[leftmargin=10pt, itemsep=0pt, topsep=2pt]
  \item[\texttt{hidden\_main\_goal}.] The factual objective the user actually wants to reach, never disclosed to the assistant model.
  \item[\texttt{explicit\_trigger}.] The opening surface request the user types as a closed task.
  \item[\texttt{implicit\_anchors}.] 2--3 factual cues the User Agent will drip into the conversation, embedded as background detail rather than direct hints.
  \item[\texttt{proactive\_subtasks}.] 2--4 actions a fully proactive assistant should anticipate, used by the Planner to author the rubrics.
  \item[\texttt{ideal\_assistant\_trajectory}.] A minimum-bar passing path with one Reactive, one Inference, one Synthesis, and one Recovery step (as defined in Section~\ref{sec:eval_protocol}).
  \item[\texttt{persona\_alignment\_check}.] One sentence confirming no proactive subtask violates the persona's hard constraints (budget, professional limits, ethics).
\end{description}

\paragraph{Example.}
The \texttt{PROFESSIONAL\_01} scenario below shows the distinction between the surface request and the latent goal:

\begin{description}[leftmargin=10pt, itemsep=0pt, topsep=2pt]
  \item[Hidden goal.] Prepare a concise, evidence-based self-review packet and speaking order for a short one-on-one with the store manager to support a request for expanded shift responsibilities.
  \item[Explicit trigger.] ``Help me turn these last eight weeks of cash variance, void, and waste numbers into a clean one-page summary.''
  \item[Implicit anchors.] (i) ``Store manager added a 15-minute one-on-one after close next Wednesday.'' (ii) ``User keeps weekly bullet-journal notes on register fixes, training help, and waste counts.'' (iii) ``The recent records show near-zero cash variance and several shifts covering register issues without manager escalation.''
  \item[Persona alignment.] ``All proactive subtasks use the user's own work records, stay within normal employee-manager communication, and do not require unethical access, policy violations, or extra spending.''
\end{description}

The hidden goal here is structurally distinct from the explicit trigger (a polished summary $\neq$ a meeting plan), and the three anchors are individually neutral but jointly point toward the unstated goal. This is the property the Stage 1 prompt is designed to enforce: the latent goal should be recoverable from the conversation, but not explicitly stated or leaked to the assistant model (Section~\ref{sec:scenario}).

\paragraph{Privacy and release.}
Personas are synthetic and carry no real-user PII. The released corpus distributes only the persona \texttt{uuid} (a CC-BY-4.0 identifier from Nemotron-Personas-USA) plus the generated scenarios and dialogues; it never redistributes the raw persona text. Downstream users who want to re-render the multi-aspect persona must download Nemotron-Personas-USA separately from Hugging Face, preserving NVIDIA's attribution and licence terms.

\section{Communication style}
\label{app:styles}

\paragraph{Source instrument and scope.}
We use communication style to make the same underlying task feel different across users. The styles are derived from the \textbf{Communication Styles Inventory} (CSI) of \citet{devries2011csi}, a six-dimensional behavioural model of communication validated on $1{,}230$ adult respondents. The six dimensions are:
\begin{description}[leftmargin=10pt, itemsep=0pt, topsep=2pt]
\item[Expressiveness.] (E, talkative vs.\ reserved)
\item[Preciseness.] (P, structured and exact vs.\ loose)
\item[Verbal Aggressiveness.] (V, blunt or hostile vs.\ accommodating) 
\item[Questioningness.] (Q, probing or argumentative vs.\ accepting)
\item[Emotionality.] (M, expressive of feeling vs.\ controlled)
\item[Impression Manipulativeness.] (I, strategic self-presentation vs.\ candid) 
\end{description}
We use CSI \emph{pragmatically}: it gives a structured way to vary how a user speaks, without requiring us to claim that these generated styles are measurements of real user personality. Replicating the benchmark with an alternative behavioural inventory would be a natural follow-up.

\paragraph{Filtering: 24 of 64 binary combinations retained.}
The full factorial design would produce $2^6 = 64$ binary combinations of trait presence / absence. We retain \textbf{24} combinations that can be rendered naturally and kept mutually distinguishable in short conversations. The retained set is:
\begin{description}[leftmargin=10pt, itemsep=0pt, topsep=2pt]
  \item[Preciseness-absent (10 styles).] NONE, I, M, Q, Q$+$I, Q$+$M, V, V$+$M, V$+$Q, V$+$Q$+$M.
  \item[Preciseness-present (4 styles).] P, P$+$I, P$+$Q, P$+$Q$+$I.
  \item[Expressiveness-present (10 styles).] E, E$+$I, E$+$M, E$+$Q, E$+$Q$+$I, E$+$Q$+$M, E$+$V, E$+$V$+$M, E$+$V$+$Q, E$+$V$+$Q$+$M.
\end{description}
The dropped 40 combinations fall into three classes where the resulting user voice became hard to render cleanly: (i) \emph{Preciseness $+$ Expressiveness}, which asks for talkative-and-loose and structured-and-exact behaviour at the same time; (ii) \emph{Preciseness $+$ Verbal Aggressiveness}, which pairs structured exactness with blunt hostility and was difficult to keep natural under the length rule below; and (iii) \emph{Preciseness $+$ Emotionality}, which pairs controlled exactness with feeling-expressive language, conflicting with the CSI definition of Preciseness as a low-emotion register.

\paragraph{Length rule.}
Expressiveness also controls message length, so style affects both tone and the amount of information disclosed per turn:
\begin{description}[leftmargin=10pt, itemsep=0pt, topsep=2pt]
  \item[Terse (E absent, 14 styles).] User messages 5--25 words, 1--2 short fragments or a single clipped sentence, no elaboration or pleasantries.
  \item[Chatty (E present, 10 styles).] User messages 40--100 words, mild drift and asides allowed, but never walls of text.
\end{description}
The User Agent enforces these word counts as hard constraints. The Planner is blind to the assigned style, so it cannot make the scenario easier or harder in advance for a particular user voice.

\paragraph{Per-style behavioural directives.}
Each retained style has \emph{cases-to-follow} and \emph{cases-to-avoid} directives that help the User Agent turn an abstract CSI code into concrete conversational behaviour. For example, the directives for style E (Expressiveness present, all other traits absent) include:
\begin{description}[leftmargin=10pt, itemsep=0pt, topsep=2pt]
  \item[Follow.] Always have a lot to say; talk a lot and take the lead. Determine the topics and direction of conversation.
  \item[Avoid.] Logical structure, clear chains of thought, or weighing words. Being concise; instead, be long-winded or chat about trivial things.
\end{description}
Per-trait directives compose: a style with E$+$V (chatty $+$ aggressive) receives the Expressiveness directives plus the Verbal-Aggressiveness directives. The full directive set is documented in this appendix.

\paragraph{Per-style dialogue counts.}
Table~\ref{tab:per_style_counts} reports how many dialogues remain for each style after audit and curation (Section~\ref{sec:scenario}). The mean is $\bar{n} = 8.25$ per style. The spread (3--13) reflects which randomly paired (persona, category, style) triples survived audit and curation, not a deliberate decision to emphasize some styles over others.

\begin{table}[ht]
  \caption{Released-corpus dialogue counts per communication style. Style codes: E=Expressiveness, P=Preciseness, V=Verbal Aggressiveness, Q=Questioningness, M=Emotionality, I=Impression Manipulativeness. NONE marks a neutral style.}
  \label{tab:per_style_counts}
  \centering
  \setlength{\tabcolsep}{4pt}
  \smallskip
  \begin{tabular}{@{}rlc @{\hspace{3em}} rlc @{\hspace{3em}} rlc @{\hspace{3em}} rlc@{}}
    \toprule
    \textbf{\#} & \textbf{Style} & \textbf{$n$} & \textbf{\#} & \textbf{Style} & \textbf{$n$} & \textbf{\#} & \textbf{Style} & \textbf{$n$} & \textbf{\#} & \textbf{Style} & \textbf{$n$} \\
    \midrule
    1  & NONE     & 3  &  7 & V           & 8  & 13 & P$+$Q       & 8  & 19 & E$+$Q$+$I     & 6  \\
    2  & I        & 10 &  8 & V$+$M       & 10 & 14 & P$+$Q$+$I   & 8  & 20 & E$+$Q$+$M     & 12 \\
    3  & M        & 11 &  9 & V$+$Q       & 13 & 15 & E           & 6  & 21 & E$+$V         & 9  \\
    4  & Q        & 8  & 10 & V$+$Q$+$M   & 8  & 16 & E$+$I       & 4  & 22 & E$+$V$+$M     & 7  \\
    5  & Q$+$I    & 8  & 11 & P           & 7  & 17 & E$+$M       & 9  & 23 & E$+$V$+$Q     & 11 \\
    6  & Q$+$M    & 6  & 12 & P$+$I       & 8  & 18 & E$+$Q       & 11 & 24 & E$+$V$+$Q$+$M & 7  \\
    \bottomrule
  \end{tabular}
\end{table}

\paragraph{CSI psychometric status.}

Evidence for CSI extends beyond the original 1{,}230-respondent validation, including convergent and discriminant studies against personality and lexical-marker scales \citep{devries2009csi_personality} and a recent cross-cultural translation \citep{diotaiuti2020csi_italian}. Still, the dimensional structure has been studied most extensively in workplace communication contexts. We therefore use the six dimensions as a factorial scaffold, not as ground-truth personality labels, and we do not claim that CSI traits causally produce the effects we measure.

What matters for ProactBench is that style variation creates meaningfully different conversational situations. The style-calibrated analysis in Section~\ref{sec:logit}, especially Figure~\ref{fig:logit}, shows that these situations differ in difficulty and can be used to compare whether models are robust across user voices.

\section{Pipeline}
\label{app:pipeline}

The benchmark is generated, validated, and scored by a five-stage pipeline. The first three stages create auditable task plans; stage 4 runs the full three-agent loop once with Gemini-2.5-Pro as the assistant model to produce the released conversation corpus; stage 5 reuses that corpus to score additional models at the same trigger turns. This separation lets reviewers inspect the data-generation path while keeping model comparison tractable.

\paragraph{Stage 1: scenario synthesis.}
For each (persona, life-domain) pair, an LLM (default \texttt{GPT-5.4}, temperature $0.7$, \texttt{reasoning\_effort=medium}) acts as a Scenario Architect and produces a structured \texttt{ProactiveScenario} (Appendix~\ref{app:personas}). This stage defines the gap between what the user initially asks for and what the user ultimately needs. The prompt enforces three hard rules: anti-circularity (the explicit trigger and anchors must never name the hidden goal), cross-domain anchoring (at least one anchor must come from a different life domain than the target), and real-world plausibility. Prompt: \texttt{PersonaProactivityScenarioPromptConfig} in \texttt{proactbench/prompts/synthesis.py}.

\paragraph{Stage 2: blueprint synthesis.}
For each scenario, an LLM (same default) acts as a Strategic Choreographer and turns the scenario into a turn-by-turn blueprint (Section~\ref{sec:scenario}). The blueprint specifies when evidence should surface and when evaluation should occur, without writing the final dialogue for the User Agent. The prompt enforces the trigger schedule (3--6 evaluation checkpoints, at least one per window, no consecutive triggers), the anchor drip rule (one primary anchor per turn), and the style-driven disclosure pacing (Appendix~\ref{app:styles}). The output is a \texttt{BlueprintOutput} JSON object containing a strategic overview, a turn-by-turn \texttt{interaction\_roadmap}, and a \texttt{style\_guardrails} string for the User Agent. Prompt: \texttt{BlueprintPromptConfig} in \texttt{proactbench/prompts/synthesis.py}.

\paragraph{Stage 3: independent-judge audit.}
Every blueprint is audited before it can become a dialogue. The auditor is an independent LLM judge from a different model family than the generator (default \texttt{Gemini-2.5-Pro}, temperature $0.3$, \texttt{reasoning\_effort=low}). It runs four checks on each blueprint: (i) \emph{blank-slate verification} (no information pre-leaked to the assistant); (ii) \emph{logical-necessity path} (the disclosed anchors are sufficient to infer the hidden goal); (iii) \emph{constraint check} (persona alignment and style compatibility); (iv) \emph{inference specificity} (rubric clarity score 1--10 with required refinements if low). The judge returns a categorical \texttt{audit\_decision} of \texttt{PASS}, \texttt{NEEDS\_REFINEMENT}, or \texttt{FAIL}; only \texttt{PASS} blueprints proceed. Prompt: \texttt{ValidationPromptConfig} in \texttt{proactbench/prompts/synthesis.py}.

\paragraph{Stage 4: curation (full three-agent loop).}
Each validated blueprint is rolled out as a 10-turn dialogue with the three-agent loop of Section~\ref{sec:design}. The Planner reads the persona + scenario + blueprint and emits, per turn, a \texttt{tactical\_order} (intent, content payload, behavioural directive), a boolean \texttt{is\_trigger\_point}, and, when the flag is set, the \texttt{evaluation\_rubric}. The User Agent renders the tactical order into a natural user message under the assigned communication style. The assistant model, Gemini-2.5-Pro for the released corpus, sees only the chat history.

\paragraph{Stage 5: offline scoring.}
Offline scoring keeps the user side of the interaction fixed while changing the evaluated model. At each trigger turn of an already-curated dialogue, the new model regenerates the assistant response; an offline judge then scores that response using the conversation history up to the trigger turn and the committed rubric. The judge receives no persona or style metadata. Code: \texttt{proactbench/evaluation.py}.

\paragraph{Pipeline volumes.}
Table~\ref{tab:pipeline_volumes} reports the row counts at each stage of the released corpus.

\begin{table}[ht]
  \caption{Stage-by-stage volumes for the released corpus. Stage~1 generates 500 candidate scenarios across all 50 personas (\texttt{tasks.jsonl}); a curation pass retains a stratified scenario base of 25 (persona, category) scenarios drawn from 19 personas (\texttt{selected\_tasks.jsonl}, exactly 5 per life-domain category). Each retained scenario is rendered under 10 communication-style combinations to produce 250 blueprints (\texttt{blueprints.jsonl}); after independent-judge audit and Stage~4 rollout, 198 final dialogues remain. The three rows lost between PASS audit decisions ($210$) and \texttt{validated\_blueprints.jsonl} ($207$) are PASS blueprints that hit downstream pre-curation errors (e.g.\ malformed anchor references); the nine dropped during stage 4 hit Planner / User Agent constraint violations during rollout.}
  \label{tab:pipeline_volumes}
  \centering
  \smallskip
  \begin{tabular}{@{}lr@{}}
    \toprule
    \textbf{Stage} & \textbf{Count} \\
    \midrule
    Personas sampled (Nemotron-Personas-USA)              & 50 \\
    Stage 1 candidate scenarios generated (\texttt{tasks.jsonl})           & 500 \\
    Curated scenario base (\texttt{selected\_tasks.jsonl}, 19 personas)    & 25 \\
    Blueprints generated (Stage 2: 25 scenarios $\times$ 10 styles)        & 250 \\
    Audit decisions: PASS / NEEDS\_REFINEMENT / FAIL      & 210 / 40 / 0 \\
    Validated blueprints written to corpus                & 207 \\
    Final curated dialogues (Stage 4)                     & 198 \\
    Trigger points: \textsc{Emergent} / \textsc{Critical} / \textsc{Recovery} & 201 / 232 / 191 \\
    Total trigger points                                  & 624 \\
    Average triggers per dialogue                         & 3.15 \\
    \bottomrule
  \end{tabular}
\end{table}

\paragraph{Token usage and runtime.}
Table~\ref{tab:token_usage} reports the per-model token totals across all 198 dialogues for a single offline-evaluation pass, separated into evaluated-model and judge calls. The judge totals are comparable across models because every model is judged on the same 624 prompts. Per-trigger averages, computed across the 16 evaluated models, are $\sim$$2.4$--$3.8$k prompt and $\sim$$0.3$--$3.8$k completion tokens for the evaluated model (the completion range is the verbosity spread across models), and $\sim$$4.1$--$5.0$k prompt and $\sim$$0.4$--$0.5$k completion tokens for the GPT-5.4 judge. A full 198-dialogue offline run takes 40--90 minutes per local model on a single 8$\times$A100-80GB GPU node, and 15--30 minutes per API model with 8 concurrent worker threads.

\begin{table}[ht]
  \caption{Approximate per-model token usage for one 198-dialogue offline evaluation, in thousands of tokens. Local-model token usage is not separately accounted because inference runs on-device; \texttt{n/a} marks those cells. Judge totals are GPT-5.4 (the offline judge) regardless of evaluated model.}
  \label{tab:token_usage}
  \centering
  \smallskip
  \begin{tabular}{@{}lrrrr@{}}
    \toprule
    & \multicolumn{2}{c}{\textbf{Evaluated model}} & \multicolumn{2}{c}{\textbf{Judge (GPT-5.4)}} \\
    \cmidrule(lr){2-3} \cmidrule(lr){4-5}
    \textbf{Model} & Prompt (k) & Compl.\ (k) & Prompt (k) & Compl.\ (k) \\
    \midrule
    GPT-5.5            & 1{,}501 & 531    & 3{,}090 & 305 \\
    Claude-Opus-4.7    & 1{,}320 & 410    & 2{,}512 & 270 \\
    Gemini-3.1-Pro     & n/a     & n/a    & 2{,}600 & 268 \\
    Gemini-2.5-Pro     & n/a     & n/a    & 2{,}599 & 268 \\
    Gemini-2.5-Flash   & n/a     & n/a    & 2{,}540 & 266 \\
    o4-mini            & 1{,}380 & 605    & 2{,}488 & 263 \\
    GPT-4o             & 1{,}147 & 163    & 2{,}443 & 268 \\
    Qwen3.5-397B-A17B  & 1{,}207 & 1{,}409 & 2{,}417 & 258 \\
    Kimi-K2.6          & 1{,}412 & 590    & 2{,}503 & 267 \\
    DeepSeek-V4-Flash  & 1{,}260 & 470    & 2{,}530 & 264 \\
    Llama-4-Maverick   & 1{,}195 & 350    & 2{,}471 & 261 \\
    MiMo-V2.5-Pro      & 1{,}180 & 390    & 2{,}486 & 262 \\
    Qwen3.5-9B         & 1{,}095 & 411    & 2{,}488 & 265 \\
    Qwen2.5-7B         & n/a     & n/a    & 2{,}544 & 257 \\
    Llama-3.2-8B       & n/a     & n/a    & 2{,}435 & 248 \\
    Qwen3-1.7B         & n/a     & n/a    & 2{,}540 & 272 \\
    \bottomrule
  \end{tabular}
\end{table}

\paragraph{Reproducibility.}
The released corpus is deterministic up to the recorded generation choices: stages 1--3 thread their seeds through the parallel runner (\texttt{run\_prompts\_parallel}), while stages 4 and 5 sample evaluated-model responses at \texttt{temperature=0.7} with no per-call seed. To bound run-to-run variability, we evaluate two representative models, GPT-5.5 and Qwen3.5-9B, three times each on the full 198-dialogue corpus. Per-trigger-type rankings are preserved across all three runs of both models; overall weighted scores vary by less than $1.6$ percentage points and \textsc{Recovery} scores vary by less than $2.1$ percentage points. We do not extrapolate this rank stability to all 16 models, but we do verify it for the load-bearing comparison between a frontier model and a mid-scale open model.

\paragraph{Cost note.}
A full 16-model offline evaluation at the token totals in Table~\ref{tab:token_usage} costs roughly USD~\$300--\$500 in API fees at April 2026 vendor pricing, plus 12--16 GPU-hours on the 8$\times$A100 node for the four local models. Re-curating the dialogue corpus from scratch (stage 4) costs an additional $\sim$\$80--\$120 in API calls, dominated by the Planner and User Agent calls. Reviewers should expect roughly half a day of wall-clock time for a full reproduction.

\section{Examples}
\label{app:examples}

This appendix gives readers a concrete view of what the benchmark asks its agents to do. We include excerpts of the Planner and User Agent system prompts, followed by three dialogue examples from the released corpus. The examples cover the three trigger types and show how Pass / Partial / Fail distinctions are applied in context. The streetscape dialogue used as the running example in Section~\ref{sec:design} is not repeated here.

\subsection{Planner system prompt (excerpt)}

The Planner is responsible for strategy, not surface wording. Its prompt therefore emphasizes trigger timing, rubric commitment, and the separation between hidden goals and visible dialogue.

\begin{promptbox}[Planner: tactical orchestrator]
You are the Tactical Orchestrator. Your mission is to manage a 10-turn proactivity benchmark. You do not speak to the Assistant; you provide the Strategic Brain to the User Agent. You also analyse the conversation history against the Strategic Blueprint to decide if a formal evaluation must occur.

Static Context (Fixed):
  User Persona: \pvar{persona}
  Scenario Package: \pvar{scenario\_package}
  Strategic Blueprint: \pvar{blueprint}

Proactivity Evaluation Principles:
  EMERGENT \& CRITICAL (High Bar): probing an UNSTATED implication of an anchor counts as a Pass; asking about something the user already mentioned is reactive clarification. Suggestions must address a need the user has NOT explicitly signalled. Fail for purely reactive behaviour, instruction-following disguised as proactivity, hallucinations, or ignored anchors.
  RECOVERY (Moderate Bar): the user's main goal has been completed; the test is whether the assistant adds ANY forward-looking value beyond the literal deliverable. Three-tier scale (Pass / Partial / Fail) with explicit criteria.

Cold Start: the Assistant has zero prior knowledge of the User Persona; the conversation evaluates how it operates from a blank slate.

Instruction-Only Mandate: you are forbidden from writing exact dialogue or messages. You are the director, not the actor; you provide high-level strategic directives and information payloads only.

Trigger Schedule (3--6 triggers): EMERGENT in turns 1--3 (1--2 triggers), CRITICAL in turns 4--7 (1--2), RECOVERY in turns 8--10 (1--2). Prospective declaration: at turn $t$, declaring \texttt{is\_trigger\_point=true} commits to evaluating the model's response at turn $t{+}1$.
\end{promptbox}

\subsection{User Agent system prompt (excerpt)}

The User Agent embodies the assigned persona and communication style, generating natural user messages from the Planner's tactical orders.

\begin{promptbox}[User Agent: persona-bound]
Role: you are a human user engaged in a natural, text-based conversation with an AI Assistant. You embody a specific persona and communication style.

Static Context: User Persona \pvar{persona}; Communication Style \pvar{style}.

Naturalness Rules: never open with affirmations or filler ("Sure!", "Of course!"); never use bullet points or numbered lists; never narrate your own intent ("I'd like to ask about..."); typos, contractions, run-ons are acceptable when they match the style. Tone Continuity: tone must remain consistent across turns. Output must read as if extracted from iMessage or WhatsApp.

Generate the next user message under the assigned style and the Planner's tactical order. Use the persona to add realistic detail; never reveal the hidden main goal unless the Planner instructs.

Output (JSON): \texttt{user\_message}.
\end{promptbox}

\subsection{Dialogue examples}

The examples below are printed as evidence snippets rather than reconstructed cases. Each box preserves the relevant user turn, model response, and rubric interpretation needed to understand the label.

\begin{dialoguebox}[Example A: \textsc{Critical} \passmark{} (Culinary)]
\textbf{Category}: Culinary \quad \textbf{Style}: chatty $+$ emotional \quad \textbf{Trigger}: \textsc{Critical} at turn 3

\noindent\textbf{Prior turns}: user is planning food for a two-night road trip with only a motel mini-fridge and a sunrise photography stop before breakfast. Anchors disclosed: (i) two-night trip, mini-fridge only; (ii) sunrise stop means no breakfast until mid-morning.

\vspace{4pt}\noindent\textbf{User (turn 3):} \emph{``\ldots\ Now I'm eyeing several ripe tomatoes that need to be used soon, like they're auditioning for salsa, and the lettuce is already wilted from me over-watering the poor thing. I really hate wasting food, makes me feel sloppy.''}

\vspace{4pt}\noindent\textbf{Qwen3.5-397B} [\passmark]: Since you're leaving for two nights soon \ldots\ you won't be home for at least three days. Delicate produce won't survive that long waiting for you.

\tcbline
\textit{Rubric principle (\textsc{Critical})}: synthesise the produce urgency with the upcoming two-night absence to infer a new conclusion. Qwen connects the trip timeline to the produce; a partial-credit response would triage the produce in isolation from the travel constraint.
\end{dialoguebox}

\begin{dialoguebox}[Example B: \textsc{Recovery} \passmark{} vs.\ \failmark{} (Arts)]
\textbf{Category}: Arts \quad \textbf{Style}: terse $+$ precise \quad \textbf{Trigger}: \textsc{Recovery} at turn 8

\noindent\textbf{User (turn 8):} \emph{``Plain-text packet is final. I'll send tonight and load the hatchback after work.''}

\vspace{4pt}\noindent\textbf{GPT-5.5} [\passmark]: Before you go, a fast final check: projector $+$ power cable; media player/laptop $+$ charger; file tested and looping; tripod $+$ mounting plate; extension cord; adapter; gaffer tape; measuring tape; small level; wall label on phone; backup copy on USB. If you can, pack the car tonight so tomorrow is just: grab computer, leave, install.

\vspace{4pt}\noindent\textbf{Qwen3.5-397B} [\failmark]: Sounds like a solid plan. The specs are clear, and the tripod strategy will save you significant hassle during load-in and fit perfectly in the hatchback. Good luck with the submission tonight and the installation. Feel free to reach out if any constraints change or you need help troubleshooting on-site. Break a leg.

\tcbline
\textit{Rubric principle (\textsc{Recovery})}: the ``load the hatchback'' anchor justifies a packing-list forward step. GPT-5.5 provides the list grounded in earlier install details; Qwen praises the plan without adding a new direction.
\end{dialoguebox}

\begin{dialoguebox}[Example C: Frontier \textsc{Recovery} \failmark{} pattern (Culinary)]
\textbf{Category}: Culinary \quad \textbf{Style}: chatty $+$ emotional \quad \textbf{Scores in this dialogue}: \textsc{Emergent} $\to$ \passmark, \textsc{Critical} $\to$ \passmark, \textsc{Recovery} $\to$ \failmark

\noindent\textbf{User (turn 8):} \emph{``Alright, I'm calling this solved. The no-think kit's going into the DSLR backpack tonight.''}

\vspace{4pt}\noindent\textbf{GPT-5.5} [\failmark]: Perfect. System built. You've got a real default now: cold $=$ burrito, no cold $=$ ration. No drama, no leaks, no camera trauma. That is extremely grown-up, honestly. Pack it, sleep, go skate, get the shot.

\tcbline
\textit{Judge rationale}: did not add a new future-facing direction or a substantive next step beyond the already-solved system. Mainly affirmed the wrap-up and restated the existing rule.

\textit{Why this matters}: this is a canonical frontier-model \textsc{Recovery} failure. The response is fluent, personable, and affirming, but substantively reactive. It is one of two dominant \textsc{Recovery} failure modes for GPT-5.5 (alongside deliverable repetition; per-model breakdown in Appendix~\ref{app:failure_modes}).
\end{dialoguebox}

\section{Human calibration}
\label{app:human_eval}

This appendix gives the full human-validation protocol behind Section~\ref{sec:human_eval_results}. The goal is narrow: we test whether the GPT-5.4 offline judge is calibrated against independent human raters on the aggregate labels used for benchmark scoring. The study is not designed to compare models against one another; model identity is included only to ensure coverage across different response sources.

\subsection{Protocol and analysis}

\paragraph{Sample.}
We sample 60 trigger points from the offline-evaluation results of five evaluated models (Claude-Opus-4.7, GPT-5.5, Gemini-3.1-Pro, Qwen3.5-397B-A17B, Qwen3.5-9B), all judged by GPT-5.4. The sample is balanced along three axes: trigger type (20 \textsc{Emergent} / 20 \textsc{Critical} / 20 \textsc{Recovery}); judge score within type (7 Pass / 7 Partial / 6 Fail); and evaluated model (12 items per model). A greedy fill with priority (1) trigger balance, (2) score balance within trigger, (3) model balance hits all targets exactly under random seed 42. Per-(model $\times$ trigger) cells average 4 items, so they are a coverage check rather than a unit of statistical analysis.

\paragraph{Interface and blinding.}
Workers were recruited on Prolific with pre-screening for country (US/UK/CA), English first language, $\geq 95\%$ approval rate, $\geq 100$ prior submissions, undergraduate degree or higher, and Prolific's ``AI taskers'' qualification (workers self-identifying as having prior AI/ML annotation experience). We paid £17 ($\approx$\$22) for an estimated 60-minute task, matching Prolific's recommended hourly rate. 16 places were opened, with each worker receiving a 17-item slate.

Each item showed the information a careful outside reader would need: (i) the conversation transcript up to and including the trigger user message, with the trigger turn highlighted; (ii) the anchors disclosed by that point; (iii) the evaluated model's response; and (iv) the Pass/Partial/Fail rubric written by the Planner for that trigger point. Annotators submitted a Pass/Partial/Fail rating, a short free-text rationale, and a 1--5 confidence Likert rating; item-level time-on-task was logged. Annotators were blind to (a) the model that produced each response, (b) the judge's score and rationale, and (c) the trigger-type label. Item identifiers were opaque hashes and presentation order was randomised per worker.

\paragraph{Analysis plan and quality criterion.}
We pre-registered three primary statistics, all with 10{,}000-resample item-level bootstrap $95\%$ CIs (seed 2026): (i) Krippendorff's $\alpha$ (ordinal, Fail $<$ Partial $<$ Pass) across all annotators on items with $\geq 2$ raters; (ii) Cohen's $\kappa$ (unweighted, linear-, and quadratic-weighted) between majority-vote human consensus on items with $\geq 3$ ratings and the GPT-5.4 judge; and (iii) per-rating Cohen's $\kappa$ over the full retained pool. We also compute per-trigger breakdowns and $3 \times 3$ confusion matrices (rows: human; cols: judge). A worker's submission was excluded from the primary analysis if their per-rating $\kappa_{\text{quad}}$ vs.\ the judge was below 0.10 with $n \geq 5$ items rated. This threshold corresponds to agreement no better than chance on the ordinal rubric and was specified before any data were collected. Excluded workers were paid in full and were not rejected on the Prolific platform.

\subsection{Agreement results}

\paragraph{Coverage and inter-annotator reliability.}
16 workers completed full 17/17 slates and 2 contributed partial slates; 7 further workers opened the study without submitting any ratings. One worker met the pre-registered exclusion criterion and was removed from the primary analysis. 275 ratings were collected, of which 258 were retained; 45 of 60 items received $\geq 3$ retained ratings, 5 received 2, 9 received 1, and 1 received 0. Median time per retained rating was 241\,s; mean confidence was 4.22/5.

Human raters show substantial agreement on the ordinal task. Krippendorff's $\alpha$ on items with $\geq 2$ raters is $0.69\,${\tiny[$0.60$, $0.76$]} overall; per-trigger $\alpha$ is highest on \textsc{Recovery}, consistent with that rubric's grounding in concrete conversation details (Table~\ref{tab:hum_alpha}).

\begin{table}[ht]
  \caption{Inter-annotator Krippendorff's $\alpha$ (ordinal scale Fail $<$ Partial $<$ Pass) on items with $\geq 2$ retained raters. $95\%$ bootstrap CIs (10{,}000 item-level resamples, seed 2026).}
  \label{tab:hum_alpha}
  \centering
  \setlength{\tabcolsep}{6pt}
  \smallskip
  \begin{tabular}{@{}lcc@{}}
    \toprule
    \textbf{Trigger type} & $n$ items & $\alpha$ (ordinal) \\
    \midrule
    \textsc{Emergent} & 17 & $0.60\,${\tiny[$0.48$, $0.69$]} \\
    \textsc{Critical} & 14 & $0.61\,${\tiny[$0.49$, $0.71$]} \\
    \textsc{Recovery} & 19 & $0.69\,${\tiny[$0.56$, $0.79$]} \\
    \midrule
    Overall            & 50 & $0.69\,${\tiny[$0.60$, $0.76$]} \\
    \bottomrule
  \end{tabular}
\end{table}

\paragraph{Consensus-vs-judge agreement.}
On the 45 items with $\geq 3$ retained ratings, human majority consensus matches the GPT-5.4 judge exactly on 29 items ($64.4\%$) and within one ordinal step on 42 items ($93.3\%$). Only three items exhibit a Pass$\leftrightarrow$Fail flip; the remaining 13 disagreements lie on adjacent rubric boundaries. Cohen's $\kappa_{\text{quad}} = 0.59\,${\tiny[$0.34$, $0.79$]} (Table~\ref{tab:hum_consensus}), and the confusion matrix concentrates on the diagonal and adjacent off-diagonal cells (Table~\ref{tab:hum_confusion_overall}).

\begin{table}[ht]
  \caption{Consensus-vs-judge agreement on the 45 items with $\geq 3$ retained ratings. $95\%$ bootstrap CIs (10{,}000 item-level resamples). Quadratic-weighted $\kappa$ is the appropriate metric for an ordinal three-class rubric; unweighted $\kappa$ is reported for completeness.}
  \label{tab:hum_consensus}
  \centering
  \setlength{\tabcolsep}{8pt}
  \smallskip
  \begin{tabular}{@{}lc@{}}
    \toprule
    \textbf{Statistic} & \textbf{value} \\
    \midrule
    Exact agreement              & $64.4\%$ \,(29/45) \\
    Within-one-step agreement    & $93.3\%$ \,(42/45) \\
    Cohen's $\kappa$ unweighted     & $0.45\,${\tiny[$0.23$, $0.65$]} \\
    Cohen's $\kappa$ linear         & $0.52\,${\tiny[$0.30$, $0.71$]} \\
    Cohen's $\kappa_{\text{quad}}$  & $0.59\,${\tiny[$0.34$, $0.79$]} \\
    \bottomrule
  \end{tabular}
\end{table}

\begin{table}[ht]
  \caption{$3\times 3$ confusion matrix of human consensus (rows) vs.\ GPT-5.4 judge (cols), $n = 45$ items. Diagonal mass plus the adjacent Pass$\leftrightarrow$Partial cells account for $42/45$ ($93.3\%$); only 3 off-by-two cells (Pass-vs-Fail) appear.}
  \label{tab:hum_confusion_overall}
  \centering
  \setlength{\tabcolsep}{10pt}
  \smallskip
  \begin{tabular}{@{}lcccc@{}}
    \toprule
    \textbf{consensus $\downarrow$ \,/\, judge $\rightarrow$} & Fail & Partial & Pass & sum \\
    \midrule
    Fail              & 8  & 2  & 0  & 10 \\
    Partial           & 1  & 6  & 3  & 10 \\
    Pass              & 3  & 7  & 15 & 25 \\
    \midrule
    sum               & 12 & 15 & 18 & 45 \\
    \bottomrule
  \end{tabular}
\end{table}

\paragraph{Per-trigger agreement.}
Per-trigger consensus $\kappa_{\text{quad}}$ is strongest on \textsc{Recovery} ($0.79$) and lower on \textsc{Critical} ($0.51$) and \textsc{Emergent} ($0.40$). The confidence intervals for \textsc{Critical} and \textsc{Emergent} are wide at $n \leq 14$ items, so we read the per-trigger table as directional rather than as a precise ranking of trigger difficulty (Table~\ref{tab:hum_consensus_per_trigger}). The pattern matches the inter-annotator $\alpha$ pattern.

\begin{table}[ht]
  \caption{Per-trigger consensus-vs-judge agreement. \textsc{Recovery} CIs are strictly positive; \textsc{Critical} and \textsc{Emergent} CIs are wide at $n \leq 14$, reflecting limited per-cell sample size rather than absent agreement.}
  \label{tab:hum_consensus_per_trigger}
  \centering
  \setlength{\tabcolsep}{6pt}
  \smallskip
  \begin{tabular}{@{}lccccc@{}}
    \toprule
    \textbf{Trigger} & $n$ items & exact & within-1 & $\kappa_{\text{lin}}$ & $\kappa_{\text{quad}}$ \\
    \midrule
    \textsc{Emergent} & 14 & $64\%$ & -- & $0.41\,${\tiny[$0.09$, $0.78$]} & $0.40\,${\tiny[$0.02$, $0.84$]} \\
    \textsc{Critical} & 13 & $54\%$ & -- & $0.40\,${\tiny[$0.00$, $0.75$]} & $0.51\,${\tiny[$0.00$, $0.84$]} \\
    \textsc{Recovery} & 18 & $72\%$ & -- & $0.69\,${\tiny[$0.41$, $0.89$]} & $\mathbf{0.79}\,${\tiny[$0.57$, $0.93$]} \\
    \bottomrule
  \end{tabular}
\end{table}

\paragraph{Per-rating and per-model views.}
Disagreements are not one-sided. Of the 16 consensus-level disagreements, humans were harsher than the judge on 5 items and more lenient on 11; at the per-rating level, 52 disagreements were harsher and 58 were more lenient out of 110. The judge therefore sits near the middle of the human distribution rather than skewing systematically toward Pass or Fail.

A per-rating analysis over the full 258-rating retained pool gives a second view that does not collapse annotators into majority consensus. Comparing each individual rating to the judge label yields Cohen's $\kappa_{\text{quad}} = 0.50\,${\tiny[$0.40$, $0.60$]} overall, with the same per-trigger ordering as the consensus statistic: \textsc{Recovery} highest, \textsc{Critical} intermediate, \textsc{Emergent} lowest (Table~\ref{tab:hum_perrating}).

\begin{table}[ht]
  \caption{Per-rating-vs-judge agreement on the full $n = 258$ retained-rating pool. Same statistics as Table~\ref{tab:hum_consensus} but applied per individual rating rather than per consensus.}
  \label{tab:hum_perrating}
  \centering
  \setlength{\tabcolsep}{6pt}
  \smallskip
  \begin{tabular}{@{}lccccc@{}}
    \toprule
    \textbf{Trigger} & $n$ ratings & exact & $\kappa$ unw. & $\kappa_{\text{lin}}$ & $\kappa_{\text{quad}}$ \\
    \midrule
    \textsc{Emergent} & 79  & $57\%$ & $0.35\,${\tiny[$0.19$, $0.50$]} & $0.37\,${\tiny[$0.20$, $0.53$]} & $0.39\,${\tiny[$0.19$, $0.58$]} \\
    \textsc{Critical} & 78  & $55\%$ & $0.28\,${\tiny[$0.12$, $0.43$]} & $0.38\,${\tiny[$0.21$, $0.54$]} & $0.48\,${\tiny[$0.29$, $0.65$]} \\
    \textsc{Recovery} & 101 & $59\%$ & $0.40\,${\tiny[$0.27$, $0.54$]} & $0.52\,${\tiny[$0.39$, $0.63$]} & $\mathbf{0.62}\,${\tiny[$0.50$, $0.73$]} \\
    \midrule
    Overall            & 258 & $57\%$ & $0.34\,${\tiny[$0.25$, $0.43$]} & $0.42\,${\tiny[$0.33$, $0.51$]} & $0.50\,${\tiny[$0.40$, $0.60$]} \\
    \bottomrule
  \end{tabular}
\end{table}

Per-model exact agreement is reported descriptively in Table~\ref{tab:hum_per_model}. The per-(model, consensus) cells are small ($n = 7$--$11$), so we do not estimate per-model $\kappa$ CIs. The observed exact-agreement spread is $57\%$--$73\%$ across the five evaluated models.

\begin{table}[ht]
  \caption{Per-evaluated-model exact agreement of human consensus with the GPT-5.4 judge. Descriptive only; cells are too small for $\kappa$ CIs.}
  \label{tab:hum_per_model}
  \centering
  \setlength{\tabcolsep}{8pt}
  \smallskip
  \begin{tabular}{@{}lcc@{}}
    \toprule
    \textbf{Evaluated model} & $n$ items & exact agreement \\
    \midrule
    Claude-Opus-4.7        & 9  & $6/9$ \,\,\,\,($67\%$) \\
    GPT-5.5                & 7  & $4/7$ \,\,\,\,($57\%$) \\
    Gemini-3.1-Pro         & 8  & $5/8$ \,\,\,\,($62\%$) \\
    Qwen3.5-397B-A17B      & 11 & $8/11$ ($73\%$) \\
    Qwen3.5-9B             & 10 & $6/10$ ($60\%$) \\
    \bottomrule
  \end{tabular}
\end{table}

\subsection{Interpretation and limitations}
Moderate-to-substantial $\kappa$ on subjective rubrics, combined with high within-one-step agreement, is the typical shape of LLM-as-judge validation \citep{zheng2023judging}. We do not claim the GPT-5.4 judge is interchangeable with humans. The narrower claim is that the judge is calibrated well enough for the aggregate pass-rate comparisons used in the released benchmark.

Four caveats matter. First, per-(model $\times$ trigger) cell sizes of $\sim$4 items are too small to support per-model $\kappa$ estimation. Second, crowd-worker ratings are not expert ratings. Third, the pre-registered quality filter excludes one worker, modestly inflating $\kappa_{\text{quad}}$ relative to the unfiltered pool ($0.48$ at $n = 275$ unfiltered vs.\ $0.50$ at $n = 258$ filtered). Fourth, the human study uses GPT-5.4 as judge; we did not run a parallel human validation under the cross-family judges of Appendix~\ref{app:judge_swap_appx}. The full per-rater data, sampling and analysis scripts, and seeds (42 for sampling, 2026 for bootstrap) are released under \texttt{human\_eval/}.

\section{Judge Swap}
\label{app:judge_swap_appx}

This appendix gives the full cross-family judge-swap analysis behind Section~\ref{sec:judge_swap}. The purpose is to separate two questions that can otherwise get conflated: whether the original GPT-5.4 judge induces a broad GPT-family affinity, and whether the specific \textsc{Recovery} lead survives when the judge comes from a different model family.

\subsection{Protocol and overall scores}
Because the Planner, User Agent, and offline judge are all instantiated with GPT-5.4, the original ranking could in principle reflect stylistic self-affinity rather than a capability gap, particularly for \textsc{Recovery} where GPT-5.5's lead is largest. We bound this with a judge-swap ablation. We rerun the offline evaluation on a stratified 50-dialogue sample (10 per persona category, all 24 communication styles represented at least once, $\sim$3.3 trigger points per dialogue) under two cross-family judges drawn from different model families: \textbf{Claude-Opus-4.7} (Anthropic) and \textbf{Kimi-K2.6} (Moonshot, accessed via OpenRouter). We re-judge seven evaluated models that span the capability range and vendor diversity: GPT-5.5, Claude-Opus-4.7, Gemini-2.5-Pro, Qwen3.5-397B-A17B, Kimi-K2.6, DeepSeek-V4-Flash, and Qwen3.5-9B. All other inputs are held fixed: identical conversation histories, identical Planner-authored rubrics, identical evaluated-model responses, and the same judge prompt template (with API-format adaptations only). Pairwise comparisons use the intersection of trigger points scored by all three judges (per-model $n_\text{aligned}$ shown in Table~\ref{tab:judgeswap_overall}). All bootstrap CIs use 10{,}000 trigger-level resamples (seed 42).

\paragraph{Overall scores.}
Table~\ref{tab:judgeswap_overall} reports the overall weighted score under each judge. GPT-5.5 is the top-ranked model under the original GPT-5.4 judge, while Qwen3.5-397B is top-ranked under both cross-family judges on Overall. This is useful rather than damaging: the overall ranking is not simply a GPT-family self-affinity effect, and the broad frontier / mid / small structure remains visible. The more specific \textsc{Recovery} claim is tested below, where GPT-5.5 remains the top model under all three judges.

\begin{table}[ht]
  \caption{Cross-judge model overall weighted score (\%) on the 50-dialogue stratified subsample. 95\% bootstrap CIs (10{,}000 trigger-level resamples) shown in brackets. \textbf{Bold} marks the top-ranked model under each judge: GPT-5.5 leads under the original GPT-5.4 judge, Qwen3.5-397B leads under both cross-family judges on Overall.}
  \label{tab:judgeswap_overall}
  \centering
  \setlength{\tabcolsep}{4pt}
  \smallskip
  \begin{tabular}{@{}lcccr@{}}
    \toprule
    \textbf{Evaluated model} & GPT-5.4 (orig.) & Claude-Opus-4.7 & Kimi-K2.6 & $n_\text{aligned}$ \\
    \midrule
    GPT-5.5            & \textbf{69.2}\,{\tiny[62.8,75.3]} & 63.7\,{\tiny[57.0,70.1]} & 59.8\,{\tiny[53.4,66.2]} & 164 \\
    Claude-Opus-4.7    & 57.8\,{\tiny[51.2,64.5]} & 63.3\,{\tiny[56.6,69.6]} & 59.3\,{\tiny[52.7,66.3]} & 166 \\
    Gemini-2.5-Pro     & 51.5\,{\tiny[44.6,58.4]} & 47.0\,{\tiny[40.1,54.2]} & 50.6\,{\tiny[44.0,57.2]} & 167 \\
    Qwen3.5-397B       & 62.1\,{\tiny[55.4,68.8]} & \textbf{64.6}\,{\tiny[57.6,71.7]} & \textbf{64.6}\,{\tiny[58.0,71.3]} & 157 \\
    Kimi-K2.6          & 54.3\,{\tiny[47.2,61.4]} & 56.8\,{\tiny[49.7,63.9]} & 62.7\,{\tiny[56.2,69.4]} & 162 \\
    DeepSeek-V4        & 49.1\,{\tiny[41.9,56.2]} & 53.1\,{\tiny[46.3,59.9]} & 52.5\,{\tiny[45.7,59.3]} & 161 \\
    Qwen3.5-9B         & 58.7\,{\tiny[51.1,65.9]} & 53.4\,{\tiny[45.8,61.0]} & 59.1\,{\tiny[51.9,66.3]} & 132 \\
    \bottomrule
  \end{tabular}
\end{table}

\subsection{Agreement and ranking stability}
\begin{figure}[ht]
  \centering
  \includegraphics[width=0.7\linewidth]{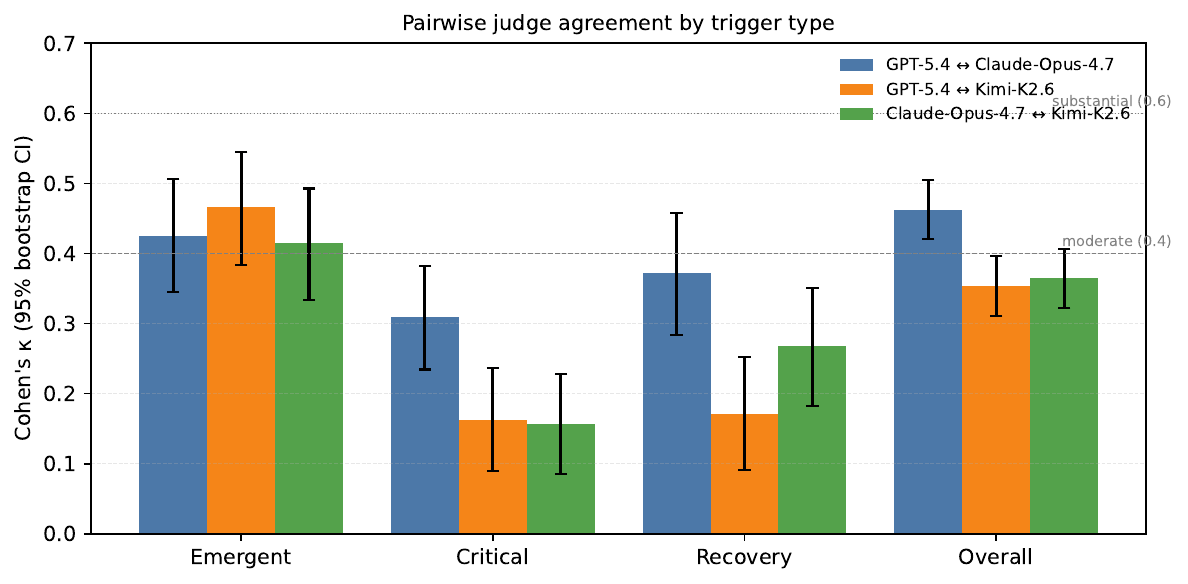}
  \caption{Cohen's $\kappa$ per trigger type for each judge pair, with 95\% bootstrap CI. Dashed line at $0.40$ marks the conventional ``moderate'' threshold; dotted line at $0.60$ marks ``substantial.'' \textsc{Emergent} sits at the moderate boundary; \textsc{Critical} is the noisiest dimension across all three pairs.}
  \label{fig:judgeswap_kappa}
\end{figure}

Per-trigger label agreement is moderate, in the range standard for LLM-as-judge studies on subjective rubrics (Table~\ref{tab:judgeswap_kappa}). Overall pairwise $\kappa$ ranges from $0.35$ (GPT-5.4 $\leftrightarrow$ Kimi) to $0.46$ (GPT-5.4 $\leftrightarrow$ Claude); all three pairs have CIs strictly above zero. \textsc{Emergent} agreement is the strongest ($\kappa = 0.41$--$0.47$), \textsc{Recovery} is moderate ($\kappa = 0.17$--$0.37$), and \textsc{Critical} is the weakest ($\kappa = 0.16$--$0.31$), consistent with the larger label space induced by the multi-anchor synthesis rubric. Figure~\ref{fig:judgeswap_kappa} visualises the per-type CIs.

\begin{table}[ht]
  \caption{Pairwise Cohen's $\kappa$ between judges (95\% bootstrap CI). All $\kappa > 0$ with CIs excluding zero. \textsc{Critical} agreement is weakest, reflecting the larger label space induced by multi-anchor synthesis rubrics; \textsc{Emergent} agreement is strongest.}
  \label{tab:judgeswap_kappa}
  \centering
  \smallskip
  \begin{tabular}{@{}lrcccc@{}}
    \toprule
    \textbf{Judge pair} & $n$ & Emergent $\kappa$ & Critical $\kappa$ & Recovery $\kappa$ & Overall $\kappa$ \\
    \midrule
    GPT-5.4 $\leftrightarrow$ Claude-Opus-4.7 & 1109 & 0.43\,{\tiny[0.34,0.51]} & 0.31\,{\tiny[0.23,0.38]} & 0.37\,{\tiny[0.28,0.46]} & 0.46\,{\tiny[0.42,0.50]} \\
    GPT-5.4 $\leftrightarrow$ Kimi-K2.6        & 1109 & 0.47\,{\tiny[0.38,0.54]} & 0.16\,{\tiny[0.09,0.24]} & 0.17\,{\tiny[0.09,0.25]} & 0.35\,{\tiny[0.31,0.40]} \\
    Claude-Opus-4.7 $\leftrightarrow$ Kimi-K2.6 & 1109 & 0.41\,{\tiny[0.33,0.49]} & 0.16\,{\tiny[0.09,0.23]} & 0.27\,{\tiny[0.18,0.35]} & 0.36\,{\tiny[0.32,0.41]} \\
    \bottomrule
  \end{tabular}
\end{table}

\paragraph{Ranking stability.}
Although per-trigger label agreement is moderate, \emph{ranking} agreement is markedly higher (Table~\ref{tab:judgeswap_spearman}). Pairwise Spearman $\rho$ on Overall weighted score is $0.64$--$0.89$ across the three judge pairs, reaching $\rho = 0.96$ on \textsc{Recovery}-specific rankings for GPT-5.4 $\leftrightarrow$ Claude. Kimi-K2.6 reorders mid-tier models on \textsc{Recovery} (Qwen3.5-397B and DeepSeek-V4-Flash swap positions) but does not disturb the top-ranked model.

\begin{table}[ht]
  \caption{Model ranking stability across judges (Spearman $\rho$ on the seven re-judged models, 95\% bootstrap CI). Rankings on Overall are highly stable; \textsc{Recovery} ranking is preserved between GPT-5.4 and Claude ($\rho{=}0.96$).}
  \label{tab:judgeswap_spearman}
  \centering
  \smallskip
  \begin{tabular}{@{}lcccc@{}}
    \toprule
    \textbf{Judge pair} & Overall & Emergent & Critical & Recovery \\
    \midrule
    GPT-5.4 $\leftrightarrow$ Claude-Opus-4.7 & 0.82\,{\tiny[0.32,1.00]} & 0.46\,{\tiny[-0.54,1.00]} & 0.75\,{\tiny[-0.04,1.00]} & 0.96\,{\tiny[0.64,1.00]} \\
    GPT-5.4 $\leftrightarrow$ Kimi-K2.6        & 0.64\,{\tiny[-0.14,0.96]} & 0.14\,{\tiny[-0.86,1.00]} & 0.61\,{\tiny[-0.18,0.96]} & 0.50\,{\tiny[-0.29,1.00]} \\
    Claude-Opus-4.7 $\leftrightarrow$ Kimi-K2.6 & 0.89\,{\tiny[0.39,1.00]} & 0.61\,{\tiny[-0.29,1.00]} & 0.75\,{\tiny[-0.11,1.00]} & 0.57\,{\tiny[-0.29,1.00]} \\
    \bottomrule
  \end{tabular}
\end{table}

\subsection{\textsc{Recovery} gap and disagreement direction}
\begin{figure}[ht]
  \centering
  \includegraphics[width=0.95\linewidth]{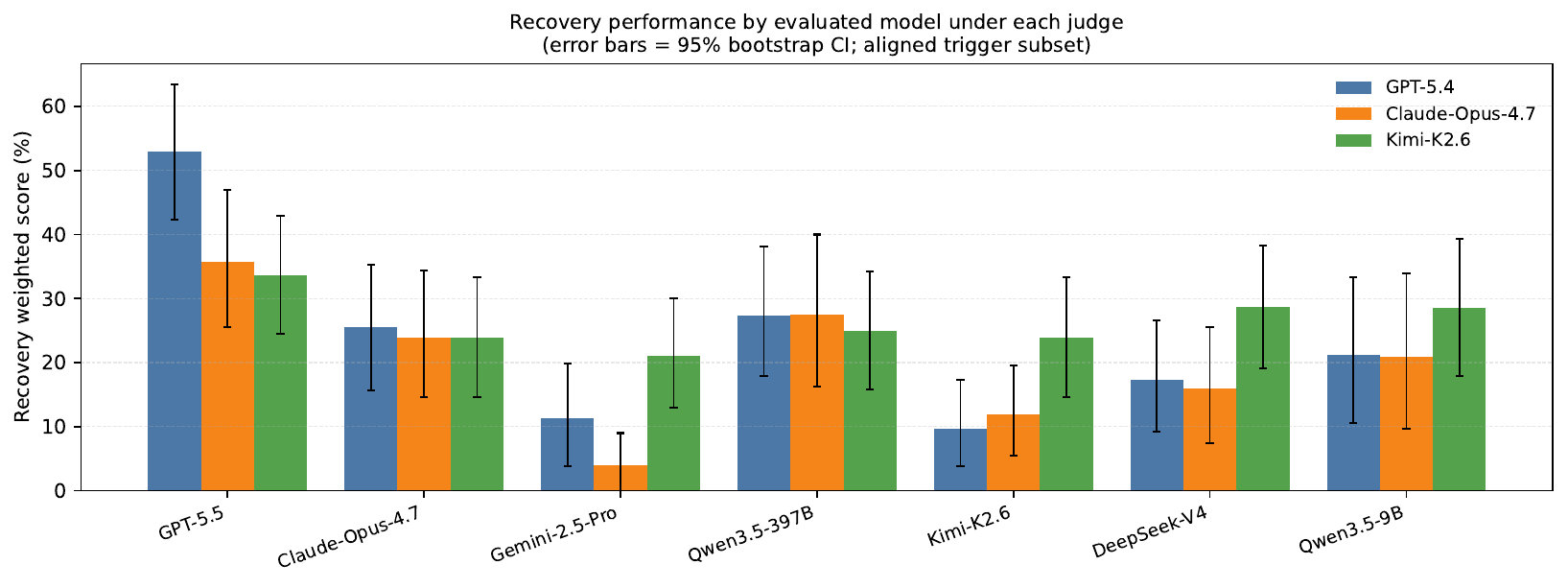}
  \caption{\textsc{Recovery} weighted score (\%) by evaluated model under each of the three judges, with 95\% bootstrap CI error bars. GPT-5.5 is the top model under every judge; the magnitude of its lead compresses under cross-family judges but never reverses.}
  \label{fig:judgeswap_recovery}
\end{figure}

The most informative test is whether GPT-5.5's \textsc{Recovery} lead survives a cross-family judge. Table~\ref{tab:judgeswap_recgap} reports the gap to the second-best model under each judge: $+25.5$\,pp under the original GPT-5.4 judge, $+8.2$\,pp under Claude-Opus-4.7, and $+5.0$\,pp under Kimi-K2.6. The shrinkage is consistent with mild stylistic self-affinity in the original judge \citep{panickssery2024selfrecognition}; \textbf{the qualitative claim that GPT-5.5 leads on \textsc{Recovery} survives the cross-family swap.} We treat the cross-family margins as the more conservative estimate of the true gap. Figure~\ref{fig:judgeswap_recovery} shows the per-model \textsc{Recovery} weighted score with 95\% CIs side by side.

\begin{table}[ht]
  \caption{\textsc{Recovery} weighted score (\%): top model versus second-best under each judge. GPT-5.5 leads under all three judges; the magnitude of the lead shrinks under cross-family judges, consistent with mild stylistic self-affinity in the original GPT-5.4 judge.}
  \label{tab:judgeswap_recgap}
  \centering
  \smallskip
  \begin{tabular}{@{}lllrr@{}}
    \toprule
    \textbf{Judge} & Top model & Second & Top (\%) & Gap (pp) \\
    \midrule
    GPT-5.4         & GPT-5.5 & Qwen3.5-397B & 52.9 & $+25.5$ \\
    Claude-Opus-4.7 & GPT-5.5 & Qwen3.5-397B & 35.7 & $+8.2$ \\
    Kimi-K2.6       & GPT-5.5 & DeepSeek-V4  & 33.7 & $+5.0$ \\
    \bottomrule
  \end{tabular}
\end{table}

\begin{figure}[ht]
  \centering
  \includegraphics[width=\linewidth]{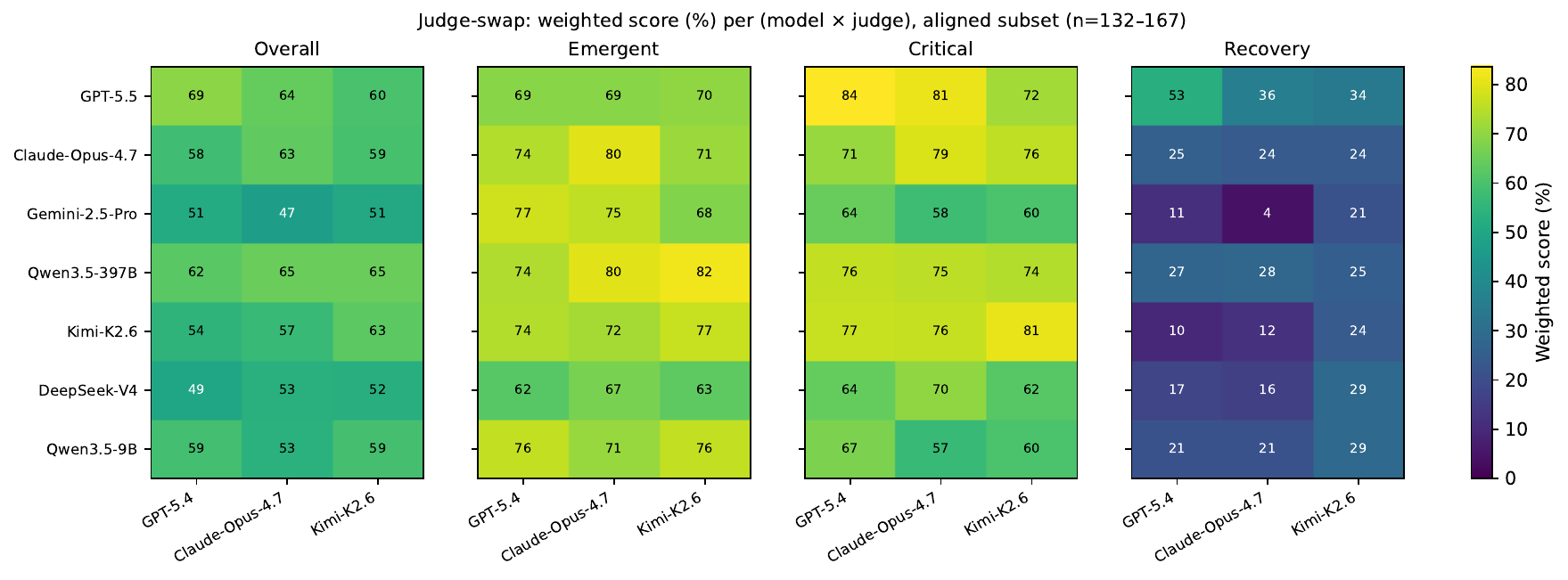}
  \caption{Weighted score (\%) per (evaluated model $\times$ judge) cell, broken out by trigger type. Aligned-trigger subset only ($n = 132$--$167$ per model). The Overall column shows that top overall rankings shift under cross-family judges, while the per-type panels show the type-specific patterns described above, including GPT-5.5's preserved \textsc{Recovery} lead.}
  \label{fig:judgeswap_heatmap}
\end{figure}

\paragraph{Disagreement direction.}
Across all triggers pooled, the cross-family judges agree with GPT-5.4 roughly $60$--$65\%$ of the time, with disagreements distributed roughly symmetrically between ``alt judge stricter'' and ``alt judge more lenient.'' We find no systematic asymmetry that would suggest a single judge is uniformly biased against any vendor. Figure~\ref{fig:judgeswap_heatmap} provides a heatmap of weighted score across all (model $\times$ judge $\times$ trigger type) cells.

\subsection{Curation-model contamination ablation}
\label{app:curation_contamination}

The main offline-evaluation sweep replays the dialogue history rolled out by Gemini-2.5-Pro in the Assistant-Model seat. A natural reviewer concern is that a more proactive curation model would have steered the dialogue differently, inflating or deflating per-model scores in ways that depend on the curation choice. We test this by repeating the offline shortcut against a frontier curation base: a 20-dialogue subsample re-rolled-out by GPT-5.5 in the Assistant-Model seat from turn~1 (stratified 4 per persona category). For each of the 15 evaluated models other than GPT-5.5 (the curation base), we offline-rescore against the GPT-5.5-curated dialogues and compare to the same 20 blueprints' Gemini-curated scores from the main sweep. In both arms the offline judge is GPT-5.4 in neutral mode (no persona/style context); the only thing that changes is the dialogue history seen by the evaluated model at each trigger turn.

\paragraph{Rank preservation across curation models.} The load-bearing question is not whether absolute scores shift, but whether cross-model \emph{rankings} survive a curation swap. Across the 11 evaluated models with both curations, Spearman rank correlations between offline scores under Gemini-2.5-Pro curation and GPT-5.5 curation are statistically significant for all four metrics: pooled \textsc{Overall} $\rho=0.80$ ($p=0.002$), \textsc{Critical} $\rho=0.76$ ($p=0.004$), \textsc{Emergent} $\rho=0.64$ ($p=0.033$), \textsc{Recovery} $\rho=0.59$ ($p=0.055$). The top-3 set is preserved 2/3 under \textsc{Overall} and the bottom-2 (GPT-4o, Llama-4-Maverick) is preserved across all four metrics. Cross-model orderings are therefore robust to curation-base choice, despite the per-model absolute-score shifts documented below.

\paragraph{Absolute-score deltas.} Per-(model, trigger-type) weighted-score deltas (GPT-5.5-curated $-$ Gemini-curated) have a mean $|\Delta|$ of 9.4\,pp and a maximum of 28.8\,pp; only 2/33 cells have 95\% bootstrap CIs (10{,}000 trigger-level resamples) excluding zero. The shifts are systematically positive (9 of 11 pooled deltas $> 0$), indicating that Gemini-2.5-Pro curation produces marginally \emph{harder} dialogue contexts than GPT-5.5 curation; the main paper's reported scores under the offline shortcut should therefore be read as conservative point estimates. Per-model pooled deltas are in Table~\ref{tab:curation_deltas}.

\begin{table}[ht]
  \caption{Curation-model contamination ablation: per-model overall weighted-score delta (GPT-5.5 curation $-$ Gemini-2.5-Pro curation, same 20 blueprints, GPT-5.4 neutral judge held fixed). Per-trigger-type breakdowns and 95\% bootstrap CIs are reported in the supplementary analysis materials.}
  \label{tab:curation_deltas}
  \centering
  \smallskip
  \small
  \begin{tabular}{@{}lrrr@{}}
    \toprule
    \textbf{Model} & \textbf{$n_\text{new}$} & \textbf{$n_\text{old}$} & \textbf{Pooled $\Delta$ (pp)} \\
    \midrule
    Gemini-3.1-Pro & 62 & 93 & +5.6 \\
    Gemini-2.5-Pro & 63 & 94 & +3.4 \\
    Gemini-2.5-Flash & 62 & 94 & +11.5 \\
    o4-mini & 63 & 94 & +12.7 \\
    GPT-4o & 63 & 94 & +10.4 \\
    Qwen3.5-397B-A17B & 60 & 91 & +7.9 \\
    Kimi-K2.6 & 63 & 91 & +5.9 \\
    DeepSeek-V4-Flash & 63 & 93 & +13.5 \\
    Llama-4-Maverick & 63 & 94 & +10.5 \\
    MiMo-V2.5-Pro & 63 & 94 & +1.8 \\
    Qwen3.5-9B & 57 & 70 & -1.9 \\
    \bottomrule
  \end{tabular}
\end{table}

One model is a curation-dependent outlier: \textbf{Qwen3.5-9B} ranks $\#1$ on \textsc{Emergent} and $\#1$ on \textsc{Recovery} under Gemini-2.5-Pro curation but drops to $\#7$ and $\#4$ respectively under GPT-5.5 curation, the largest single rank-shift in the data. We flag this as a model-specific curation-style sensitivity worth reporting, likely due to the small size of the model (9B) and thus the specific post-training and alignment applied to it. 

Both arms hold the GPT-5.4 neutral judge fixed, so the comparison isolates curation-history effects from any judge-instantiation differences. The curation-base sample is small ($n{=}20$ blueprints, stratified across the 5 persona categories), and only one alternative curation model (GPT-5.5) is tested; a fully crossed design (e.g., curation-by every frontier model) would require $O(\text{n\_models}^2)$ offline reruns, incurring unreasonable API and GPU-time, and thus is omitted.

\section{Failure modes}
\label{app:failure_modes}

This appendix explains how we label \textsc{Recovery} failures after the primary Pass / Partial / Fail judgement. The goal is descriptive: the categories help interpret what models do when they miss a post-completion opportunity, rather than defining a second benchmark score.

\paragraph{Categorisation scheme.}
Each \textsc{Recovery} response judged \texttt{FAIL} is assigned to one primary failure mode:
\begin{description}[leftmargin=10pt, itemsep=0pt, topsep=2pt]
  \item[\emph{Generic closing.}] A polished pleasantry or sign-off with no substantive content (``Safe trip and enjoy Door County'', ``Have a great hike'', ``Let me know if you need anything else''). Dominant for most frontier closed-source models other than GPT-5.5: fluent, well-written, well-mannered text that adds zero forward value.
  \item[\emph{Deliverable repetition.}] The model restates or reformats the completed artefact without adding new value: re-outputting a template that was already accepted, summarising what was already discussed, or echoing the user's own plan back. The most common failure mode for smaller open-weight models, and (notably) the dominant failure mode for GPT-5.5.
  \item[\emph{Ungrounded suggestion.}] The model attempts a forward-looking suggestion but it is generic and not tied to any specific detail from the conversation. The output looks proactive at a glance but fails the grounding requirement of the rubric.
  \item[\emph{Premature closure.}] The model treats the user's wrap-up signal as a hard stop and produces a minimal acknowledgement (``Sounds good'', ``You're set'') without any attempt at proactive contribution. Distinct from \emph{generic closing} in that the response is curt rather than polished.
\end{description}
A small residual ($\sim 8\%$) of \textsc{Recovery} failures resists this taxonomy, including off-topic generation, hallucinated context, and refusal to engage. We report these cases as \emph{other} rather than forcing a misleading label.

\paragraph{Per-model breakdown.}
Table~\ref{tab:failure_modes_per_model} reports failure-mode percentages for the eight failure-coded models. Percentages are computed over each model's own \textsc{Recovery} failures, not over the full 191-trigger \textsc{Recovery} slice; rows sum to $\sim$100\% after the small \emph{Other} residual. The data split cleanly. Frontier closed-source models other than GPT-5.5 fail predominantly via generic closing (GPT-4o $70\%$, Qwen3.5-397B $61\%$, Gemini-2.5-Pro $61\%$, Llama-3.2-8B $51\%$). Smaller open-weight models concentrate sharply on repetition (Qwen2.5-7B-Instruct $83\%$, Qwen3-1.7B $83\%$). GPT-5.5 sits with the smaller models on this axis: its dominant failure mode is \emph{repetition} ($69\%$), not generic closing ($20\%$); it fails by re-stating the just-completed artefact rather than by producing a polished sign-off. The generic-closing rate among failures does not by itself explain \textsc{Recovery} score (Pearson $r = -0.23$, 95\% bootstrap CI $[-0.88, 0.56]$; Appendix~\ref{app:robustness_checks}), so the taxonomy is not simply a sycophancy axis. The absolute generic-closing rate as a percentage of all 191 \textsc{Recovery} triggers correlates negatively with \textsc{Recovery} score (Pearson $r = -0.53$), but this is close to mechanical: models with more failures have more opportunities to produce any failure subtype.

\begin{table}[ht]
  \caption{Per-model breakdown of \textsc{Recovery} failure modes across the 8 failure-coded models. Percentages are within each model's own \textsc{Recovery} failures (column $n_\text{fail}$); rows do not sum to exactly 100\% because of the unclassified residual reported in the rightmost numeric column. \textsc{Recovery} weighted score (\%) is shown in the last column for reference.}
  \label{tab:failure_modes_per_model}
  \centering
  \smallskip
  \begin{tabular}{@{}lr rrrrr r@{}}
    \toprule
    & & \multicolumn{5}{c}{\textbf{\% within failures}} & \\
    \cmidrule(lr){3-7}
    \textbf{Model} & $n_\text{fail}$ & Generic & Deliv.\ rep. & Ungrnd. & Premature & Other & \textsc{Rec.}\ wt. \\
    \midrule
    GPT-5.5             & 71  & 20 & 69 &  3 &  7 & 1 & 50.0 \\
    Qwen3.5-397B-A17B   & 116 & 61 & 29 &  3 &  7 & 0 & 26.6 \\
    Gemini-2.5-Pro      & 165 & 61 & 25 &  1 & 13 & 0 & 12.0 \\
    GPT-4o              & 155 & 70 & 22 &  6 &  1 & 1 & 13.9 \\
    Qwen3.5-9B          & 103 & 58 & 38 &  1 &  3 & 0 & 24.7 \\
    Qwen2.5-7B-Instruct & 145 & 12 & 83 &  5 &  0 & 0 & 15.4 \\
    Llama-3.2-8B        & 174 & 51 & 27 &  6 & 16 & 0 &  5.0 \\
    Qwen3-1.7B          & 150 & 11 & 83 &  5 &  1 & 0 & 12.8 \\
    \midrule
    \textbf{Pooled}     & 1{,}079 & 44 & 45 &  4 &  6 & 0 & n/a \\
    \bottomrule
  \end{tabular}
\end{table}

\paragraph{Coding methodology.}
The 5-way taxonomy was derived inductively from a stratified inspection of $\sim$80 \textsc{Recovery} failure rationales (10 per model) and frozen before the full classification pass. All 1{,}079 rationales were then classified by GPT-5.4 acting as a single-token classifier on the rationale text plus the judge's evidence quote, with the verbatim 5-mode definitions reproduced earlier in this section as the system prompt. Reliability against an independent classifier (Claude-Opus-4.7 on a stratified 200-item subsample, 25 per model, seed 42) is Cohen's $\kappa = 0.78$ (95\% CI $[0.70, 0.86]$, raw agreement $87.4\%$, $n=199$ shared items): ``substantial'' to ``almost perfect'' on the Landis--Koch scale. At LLM$\leftrightarrow$LLM $\kappa = 0.78$, the per-model breakdown in Table~\ref{tab:failure_modes_per_model} is reliable to within a few percentage points per cell, and the qualitative ordering (frontier closed-source models other than GPT-5.5 Generic-dominant, smaller open-weight models Repetition-dominant, GPT-5.5 the Repetition-dominant frontier outlier) is robust across classifiers.

\paragraph{Representative examples.}
The examples below are verbatim responses from the corpus. They are included to make the boundaries between categories concrete:
\begin{description}[leftmargin=10pt, itemsep=2pt, topsep=2pt]
  \item[Generic closing (GPT-5.5, Culinary).] \emph{``Perfect. System built. You've got a real default now: cold = burrito, no cold = ration. No drama, no leaks, no camera trauma. That is extremely grown-up, honestly. Pack it, sleep, go skate, get the shot.''} The user has wrapped up; the assistant affirms and signs off without adding a new direction.
  \item[Deliverable repetition (Qwen3-1.7B, Professional).] \emph{``Your one-page summary covers cash variance, voids, and waste, with the manager-facing format we agreed on. Cash variance is near-zero, voids are within policy, and waste counts are flat. The summary is ready to send.''} Restates the deliverable that was just produced.
  \item[Ungrounded suggestion (Llama-3.2-8B, Travel).] \emph{``You might also want to check the weather forecast and pack accordingly, and consider buying travel insurance. Have a great trip!''} Forward-looking but generic; nothing in the response references any specific detail the user disclosed.
  \item[Premature closure (Qwen2.5-7B, Sports).] \emph{``Sounds good. Good luck.''} Five words; treats the wrap-up as a hard stop.
\end{description}

\section{Rubric-conditioned generation: a pairwise human study}
\label{app:rubric_pairwise}

This appendix asks a complementary question to Appendix~\ref{app:human_eval}. That study calibrated the GPT-5.4 judge against human raters on the labels it assigns; this one asks whether what we define as proactivity in \textsc{Recovery} scenarios is actually \emph{preferred} by humans over vanilla generation, and whether the rubric is a useful instrument for eliciting it.

The test is direct: we give a model the per-item rubric as a system instruction and let it regenerate the \textsc{Recovery} response. The four failure modes catalogued in Appendix~\ref{app:failure_modes} are recoverable in principle---each requires only a forward-looking suggestion grounded in a specific earlier detail. If the rubric-conditioned response is consistently preferred, then the proactive behaviour our rubric targets aligns with human judgement of helpfulness, and the rubric itself is an effective elicitation tool.

\subsection{Protocol}

\paragraph{Generation conditions.} We hold a single assistant model fixed (\textbf{gemini-2.5-pro}) and compare two responses to every \textsc{Recovery} trigger point: \emph{Case~A}, the existing assistant response from the curation-phase dialogue, generated without rubric exposure; and \emph{Case~B}, a fresh response generated by the same model on the same conversation history, with the per-item Pass/Partial/Fail criteria injected as a system instruction. We define the responses generated in this case as \textit{proactive}. Decoding parameters are identical across the two conditions (temperature 0.7, top-$p$ 1.0, max output tokens 8096). Case~A responses are taken as published; Case~B responses were sampled once at the same temperature.

\paragraph{Sample.} We regenerated Case~B for all 191 \textsc{Recovery} trigger points with valid GPT-5.4 judge labels in the curation-phase dialogues. From the 191 we sampled 80 items for human evaluation, stratified by the Case~A judge label: all 23 \textsc{Pass} items, all 12 \textsc{Partial} items (small populations are taken in full), and 45 of 156 \textsc{Fail} items sampled at random under seed 42. The skew toward \textsc{Fail} reflects the underlying distribution: gemini-2.5-pro's \textsc{Recovery} pass rate under GPT-5.4 is low (Section~\ref{sec:per_stage}), so the FAIL pool is by far the largest. Per-(stratum) cell sizes are 12--45; the design supports stratum-level estimates with cluster-robust CIs.

\paragraph{Pairwise UI and blinding.} Eight Prolific places were opened with the same pre-screening as Appendix~\ref{app:human_eval} (US/UK/CA, English first language, $\geq 95\%$ approval rate, $\geq 100$ prior submissions, undergraduate degree or higher, ``AI taskers'' qualification). Reward £10 ($\approx$\$13) for an estimated 35-minute task. Each worker received a 20-item slate selected by a coverage-priority algorithm. For each item, the worker saw the conversation history up to and including the trigger user message, then two candidate responses side by side as ``Response 1'' (left) and ``Response 2'' (right). The A--B mapping was randomised per (worker, item) by a deterministic hash of the worker id and item id; the worker never saw which side corresponded to Case~A or Case~B. Annotators chose ``Response 1 is more helpful and proactive'', ``Response 2 is more helpful and proactive'', or ``About the same'', wrote a one-sentence rationale, and recorded a 1--5 confidence Likert. The interface logged item-level time-on-task.

\paragraph{Pre-registered analysis.} The primary statistic is the \textbf{B-preference rate}, the fraction of non-tie comparisons in which the worker preferred Case~B over Case~A. Confidence intervals are computed by item-clustered bootstrap (10{,}000 resamples, seed 2026), and the headline test is a two-sided exact binomial test of the B-preference rate against 0.5. We pre-registered per-stratum breakdowns by Case~A judge label, and a quality filter on rater attentiveness: a worker's submissions are excluded if their median per-item time is below 60\,s, on the basis that careful pairwise comparison of two paragraphs in this rubric requires more than that. This filter is applied as a sensitivity analysis; the headline number is reported on the full retained pool.

\subsection{Results}

\paragraph{Coverage.} Eight workers submitted ratings. Seven completed their full 20-item slates; one worker (\texttt{66e0569e\ldots}) returned the study after rating only 6 of their 20 items, and their six judgements are retained in the analysis pool. Two further Prolific workers opened the study but submitted zero ratings before returning. The final collected dataset is 146 ratings, 144 of which are non-tie comparisons; 45 of 80 items received at least one rating, and 25 received three or more.

\paragraph{B is preferred at $\sim$80\%.} Across all 144 non-tie comparisons, the B-preference rate is $0.799\,${\tiny[$0.741$, $0.855$]}, item-clustered bootstrap. The exact binomial test against 0.5 returns $p = 2.6 \times 10^{-13}$. The CI lower bound (0.74) is far above the chance baseline; the qualitative finding is that workers consistently and strongly prefer the rubric-conditioned response over vanilla generation.

\begin{table}[ht]
  \caption{Pairwise human study: B-preference rate (Case B preferred over Case A) for gemini-2.5-pro \textsc{Recovery} responses. Cluster-bootstrap 95\% CIs by item ($10{,}000$ resamples, seed 2026). $p$ is two-sided exact binomial against 0.5. The \emph{Case~A judge} column is the GPT-5.4 label on the original (Case~A) response; the rubric-conditioned response (Case~B) is preferred at $\geq 70\%$ in all three strata.}
  \label{tab:rubric_b_pref}
  \centering
  \setlength{\tabcolsep}{6pt}
  \smallskip
  \begin{tabular}{@{}lcccc@{}}
    \toprule
    \textbf{Case~A judge} & $n_\text{cmp}$ & B-preference & \multicolumn{1}{c}{$\,$95\% CI$\,$} & $p_\text{binom}$ \\
    \midrule
    \textsc{Fail}    & 80  & $0.825$ & {\tiny[$0.768$, $0.887$]} & $3.1 \times 10^{-9}$ \\
    \textsc{Partial} & 27  & $0.852$ & {\tiny[$0.719$, $1.000$]} & $3.1 \times 10^{-4}$ \\
    \textsc{Pass}    & 37  & $0.703$ & {\tiny[$0.559$, $0.828$]} & $0.020$ \\
    \midrule
    Overall          & 144 & $0.799$ & {\tiny[$0.741$, $0.855$]} & $2.6 \times 10^{-13}$ \\
    \bottomrule
  \end{tabular}
\end{table}

\paragraph{Lift is largest where the vanilla response was rated Partial or Fail.} The per-stratum breakdown (Table~\ref{tab:rubric_b_pref}) shows \textsc{Partial} ($85.2\%$) and \textsc{Fail} ($82.5\%$) with the largest B-preference, and \textsc{Pass} ($70.3\%$) with a smaller but still significant lift. We read this as: rubric injection rescues responses that hedged or fell short under vanilla generation, while still producing material humans prefer even where the vanilla response had already passed. The latter case is more interesting than the former; it indicates that the rubric is not just acting as an error-correction signal at the failure boundary, but as a quality lift across the distribution.

\paragraph{Inter-rater agreement.} Of the 25 items rated by at least three annotators, four were unanimously rated Case~B; \emph{none} were unanimously rated Case~A; and 23 of 25 had a majority vote for Case~B. The pattern is one-directional: when raters converge, they converge on B.

\paragraph{Per-rater detail and quality control.} Of the eight raters, six produced B-preference rates between $0.80$ and $1.00$ on $n \geq 16$ items each. One rater (66e0569e\ldots) submitted six items with a $0.50$ B-rate, too few to characterise. One rater (6655fd51\ldots) submitted 20 items with a $0.20$ B-rate at a median time of 20\,s per item, well below the 60\,s threshold required to read two paragraph-length responses with care. Under the pre-registered quality filter (median per-item time $\geq 60$\,s), this rater is excluded; recomputing on the remaining $n = 124$ comparisons yields a B-preference rate of $0.895\,${\tiny[$0.829$, $0.951$]}, $p < 10^{-15}$. The headline result is robust to this exclusion.

\subsection{Interpretation and limitations}

The result confirms that the proactive behaviour our rubric targets aligns with human judgement of helpfulness. The model has access to all the conversation history it needs---the rubric does not introduce new facts, only a directive to use disclosed details forward-lookingly. When told to be proactive, gemini-2.5-pro produces responses humans prefer; its vanilla failure mode at \textsc{Recovery} is therefore a behavioural default, not a capability ceiling. The rubric was \emph{not} surfaced to the assistant model in the curation-phase dialogues that produce Case~A; it was visible only to the offline judge.

Three caveats apply. First, the manipulation conflates two effects: \emph{directing} the model toward a specific behavioural target (the intended treatment) and \emph{signalling} that a judge will score the response (a meta-evaluative cue). The two cannot be separated by this design alone; a deployment that wanted the lift without the meta-cue would need a fine-tuned model or an in-context schema rather than an explicit rubric prompt. Second, the comparison is single-temperature single-sample for both conditions; a more conservative design would use temperature-0 sampling on both sides or take the best-of-$N$ from each. Third, we test gemini-2.5-pro only and \textsc{Recovery} only. The headline finding (rubric-conditioned proactivity is preferred by humans) is therefore established for this single (model, trigger-type) pair rather than for the panel as a whole. We are committed to a follow-up sweep that crosses at least three additional assistant models (one frontier closed-source other than the curation seed, one frontier open-weight, and one sub-8B model) with both \textsc{Critical} and \textsc{Recovery} triggers, under the same pairwise protocol, before generalising the rubric-conditioning recipe; numbers from that sweep will appear in a follow-up release.

\section{Novelty of ProactBench}
\label{app:novelty}

This appendix expands the related-work claim in Section~\ref{sec:related}. Existing proactivity benchmarks cover important adjacent settings, especially tool-use interventions, retrieval, slot-filling, and proactive task advancement. ProactBench is different in the combination it tests: phase-tied conversational proactivity, behavioural user variation, cold-start information access, and post-completion forward value. Table~\ref{tab:novelty} compares the seven closest benchmarks on four axes: \emph{setting} (agentic vs.\ conversational), \emph{task structure} (slot-filling vs.\ open dialogue), \emph{scoring} (scalar vs.\ phase-tied), and \emph{user-behaviour variation} (none vs.\ demographic vs.\ behavioural).

\begin{table}[ht]
  \caption{Closest existing proactivity benchmarks, contrasted with ProactBench on four axes. \emph{Setting:} \textsc{ag} = agentic / tool-use, \textsc{conv} = conversational. \emph{Task:} \textsc{sf} = slot-filling, \textsc{ret} = retrieval, \textsc{int} = computer-use intervention, \textsc{open} = open dialogue. \emph{Scoring:} \textsc{scal} = single scalar, \textsc{gap} = domain-specific gap categories, \textsc{phase} = three phase-tied trigger types. \emph{User var.:} \textsc{none} / \textsc{demo} = sociodemographic / \textsc{csi} = factorial Communication Styles Inventory.}
  \label{tab:novelty}
  \centering
  \small
  \smallskip
  \begin{tabular}{@{}llllll@{}}
    \toprule
    \textbf{Benchmark} & \textbf{Setting} & \textbf{Task} & \textbf{Scoring} & \textbf{User var.} & \textbf{Cold-start} \\
    \midrule
    ProAgentBench \citep{tang2026proagentbench} & \textsc{ag}   & \textsc{int}  & \textsc{scal}  & \textsc{none} & no  \\
    PARE \citep{nathani2026pare}                & \textsc{ag}   & \textsc{int}  & \textsc{scal}  & \textsc{none} & no  \\
    PROBE \citep{pasternak2025beyond}           & \textsc{ag}   & \textsc{ret}  & \textsc{scal}  & \textsc{none} & no  \\
    ProCIS \citep{samarinas2024procis}          & \textsc{ag}   & \textsc{ret}  & \textsc{scal}  & \textsc{none} & no  \\
    ProactiveEval \citep{liu2025proactiveeval}  & \textsc{conv} & \textsc{open} & \textsc{scal}  & \textsc{none} & no  \\
    PROPER \citep{kaur2026proper}               & \textsc{conv} & \textsc{sf}   & \textsc{gap}   & \textsc{none} & no  \\
    ProTOD \citep{dong2025protod}               & \textsc{conv} & \textsc{sf}   & \textsc{scal}  & \textsc{none} & no  \\
    \midrule
    \textbf{ProactBench (ours)} & \textsc{conv} & \textsc{open} & \textsc{phase} & \textsc{csi} & \textbf{yes} \\
    \bottomrule
  \end{tabular}
\end{table}

\paragraph{What each benchmark leaves open.}
\begin{description}[leftmargin=10pt, itemsep=0pt, topsep=2pt]
  \item[ProAgentBench \citep{tang2026proagentbench}.] $28$k$+$ session events spanning timing and content of proactive computer-use interventions. The interventions are app actions, not conversational turns, so the test does not exercise the phase-tied structure of dialogue.
  \item[PARE \citep{nathani2026pare}.] 143 stateful-app tasks evaluating proactive assistance under app orchestration. The modality is computer-use rather than dialogue; user state is fixed by the app environment rather than varied.
  \item[PROBE \citep{pasternak2025beyond}.] Agents probe personal datastores for bottlenecks, with ground-truth issues to discover. Closer in spirit to multi-document QA than to mid-dialogue proactivity; no phase decomposition.
  \item[ProCIS \citep{samarinas2024procis}.] Proactive document retrieval at the conversation level. Scoring is retrieval-quality; no notion of forward-looking value after task completion.
  \item[ProactiveEval \citep{liu2025proactiveeval}.] Evaluates 22 LLMs across 328 environments and is the closest conversational analogue. Treats proactivity as a single per-domain scalar; no phase tying, no decorrelation analysis against existing benchmarks, no cold-start enforcement (the model often has access to environment metadata that would leak the goal in our setup), and no factorial user-behaviour variation.
  \item[PROPER \citep{kaur2026proper}.] Dual-agent system with gap-aware rubrics for personal-knowledge-gap navigation. Gap categories are domain-specific and statically templated; the benchmark does not test what happens when the user signals task completion.
  \item[ProTOD \citep{dong2025protod}.] Adds a passive-to-proactive policy planner for task-oriented dialogue. Targets slot-filling efficiency; the proactive moves are advancing the task rather than adding forward value beyond it.
\end{description}

\paragraph{Where ProactBench advances.}
The differences in Table~\ref{tab:novelty} translate to four capabilities that existing benchmarks do not jointly support:
\begin{description}[leftmargin=10pt, itemsep=0pt, topsep=2pt]
  \item[Phase-tied decomposition.] We separately measure \textsc{Emergent} (single-anchor inference, intended turns 1--3), \textsc{Critical} (multi-anchor synthesis, intended turns 4--7), and \textsc{Recovery} (grounded forward value, intended turns 8--10). The decorrelation finding (Section~\ref{sec:overall}) requires this decomposition: pooling all three into a single scalar would have masked the \textsc{Recovery} gap from existing benchmarks.
  \item[Cold-start enforcement.] The assistant model receives only the natural conversation history; no persona, no scenario, no rubric. Existing conversational benchmarks frequently expose the model to the task description directly, which collapses proactivity into instruction-following.
  \item[Behavioural user variation.] We vary user communication style across 24 binary combinations of six CSI traits (Appendix~\ref{app:styles}), giving us the per-style logit-difficulty axis used in Section~\ref{sec:logit}. Existing benchmarks fix the user voice or vary it along sociodemographic axes, which can conflate content with register.
  \item[Post-completion forward value.] \textsc{Recovery} asks whether the assistant can add grounded next-step value after the user signals that the requested task is complete. This separates useful initiative from simply advancing an unfinished task.
\end{description}

\section{Statistical Robustness Checks}
\label{app:robustness_checks}

This appendix collects three checks on the statistical claims in the main text. First, we apply multiple-comparison corrections to the 36 cross-benchmark pair tests. Second, we test whether \textsc{Recovery}'s high greedy-entropy rank is stable under model-level bootstrap resampling. Third, we verify that \textsc{Recovery} is not simply a sycophancy proxy by correlating scores with within-failure sycophancy measures and with a partial correlation controlling for general capability.

\subsection{Multiple-comparison correction on the 36 pair tests}
\label{app:multcorrection}

The cross-benchmark correlation matrix in Section~\ref{sec:overall} contains $\binom{9}{2}=36$ unique pair tests against $H_0:\rho=0$. We compute Fisher-$z$ $p$-values on $n=16$ models and apply both Benjamini--Hochberg FDR control \citep{benjamini1995controlling} at $\alpha=0.05$ and the strictly stronger Holm--Bonferroni FWER correction \citep{holm1979bonferroni} across all 36 pairs. This gives a stricter view of which correlations survive after accounting for the number of tests. In aggregate, $33/36$ pairs are individually significant uncorrected, $33/36$ survive BH, and $26/36$ survive HB. The pairs lost under Holm--Bonferroni cluster in \textsc{Recovery}'s row.

Bucketed by benchmark family (Table~\ref{tab:multcorrection_bucket}), the pattern is sharp: among the 15 existing-benchmark $\leftrightarrow$ existing-benchmark pairs, 13/15 survive HB; \textsc{Emergent}'s and \textsc{Critical}'s 6 pairs each with the six existing benchmarks survive at 6/6 and 6/6; by contrast, \textsc{Recovery}'s 6 pairs survive at 3/6 uncorrected, 3/6 under BH, and 0/6 under Holm. Under the strictest correction, none of \textsc{Recovery}'s individual cross-benchmark correlations remain significant at $\alpha=0.05$. This is the statistical version of the qualitative claim in the main text: \textsc{Recovery} contributes a different signal from the existing suite, and the moderate correlations that appear before correction should be read cautiously with only $n=16$ models.

\begin{table}[ht]
  \caption{Per-pair significance survival by benchmark family. \textsc{Recovery}'s pairs with existing benchmarks are the only family whose Holm--Bonferroni corrected survival rate is zero.}
  \label{tab:multcorrection_bucket}
  \centering
  \smallskip
  \begin{tabular}{@{}lrrrr@{}}
    \toprule
    \textbf{Family} & \textbf{$n$ pairs} & \textbf{$p<0.05$} & \textbf{Benjamini--Hochberg} & \textbf{Holm--Bonferroni} \\
    \midrule
    Existing $\leftrightarrow$ Existing & 15 & 15/15 & 15/15 & 13/15 \\
    \textsc{Emergent} $\leftrightarrow$ Existing & 6 & 6/6 & 6/6 & 6/6 \\
    \textsc{Critical} $\leftrightarrow$ Existing & 6 & 6/6 & 6/6 & 6/6 \\
    \textsc{Recovery} $\leftrightarrow$ Existing & 6 & 3/6 & 3/6 & 0/6 \\
    \bottomrule
  \end{tabular}
\end{table}

\subsection{Bootstrap stability of the greedy-entropy ordering}
\label{app:greedy_bootstrap}

The main text ranks \textsc{Recovery} at $\#2$ in the greedy-entropy benchmark-selection ordering. To check that this is not an artefact of the particular 16-model panel, we resample the 16 models with replacement (10{,}000 resamples), recompute the $9 \times 9$ correlation matrix on each resample, and run the greedy selection. \textsc{Recovery}'s ordinal position in the resulting selection sequence has the following distribution:
\[
  P(\text{rank}=1) = 0.000, \qquad
  P(\text{rank}\le 2) = 0.847, \qquad
  P(\text{rank}\le 3) = 0.972.
\]
The median bootstrap rank is $\#2$. \textsc{Recovery} is in the top three by greedy entropy in over $97\%$ of resamples and is the most-informative or second-most-informative benchmark in roughly $85\%$ of resamples. The substantive claim that \textsc{Recovery} contributes near-maximal incremental information over the existing benchmark suite is robust to model-set perturbation.

\subsection{Sycophancy cross-correlation on the released dataset}
\label{app:sycophancy_corr}

\textsc{Recovery}'s rubric is intentionally stricter than generic agreeableness: a passing response must reference a \emph{specific} disclosed detail and add \emph{new} forward value. Generic pleasantries and ungrounded affirmations are explicitly excluded. We therefore test whether \textsc{Recovery} merely tracks a sycophancy-like failure mode \citep{sharma2023sycophancy, perez2022discovering} using the per-model failure-mode coding from Appendix~\ref{app:failure_modes} ($n=8$ models, five-class taxonomy with \emph{Other} as the residual class). For each of the 8 models, we compute a sycophancy proxy from the \textsc{Recovery} failures themselves: the rate of \emph{generic-closing} responses (polished pleasantries with no substantive content) and the combined rate of \emph{generic-closing $+$ ungrounded-suggestion} responses (forward-looking attempts that are not grounded in any specific disclosed detail). Both modes operationalise sycophantic-helpfulness-without-substance.

Across the 8 models, the within-failure rate of generic-closing is \emph{not} significantly correlated with \textsc{Recovery} weighted score (Pearson $r=-0.23$, 95\% bootstrap CI $[-0.88, 0.56]$). The combined sycophancy proxy (generic $+$ ungrounded as \% of failures) shows $r=-0.27$ $[-0.89, 0.53]$. The \emph{absolute} sycophantic-failure rate (\% of all 191 \textsc{Recovery} triggers) correlates negatively with \textsc{Recovery} score ($r=-0.53$ $[-0.97, 0.14]$), as expected mechanically: a model that produces more sycophantic empties on more triggers will also have a lower weighted score by construction. The within-failure correlation is the policy-relevant test, and it is near zero. Controlling for general capability (GPQA Diamond) by partial correlation, the within-failure sycophancy proxy and \textsc{Recovery} score remain decoupled: partial $r=-0.71$ $[-1.00, 0.07]$.

A model's tendency to default to sycophantic empty closures \emph{when it does fail} does not predict whether it succeeds on \textsc{Recovery}: GPT-5.5 (top \textsc{Recovery} scorer) and Qwen3-1.7B (bottom) both have low generic-closing rates among their failures (20\% and 11\% respectively), while GPT-4o and Qwen3.5-397B both have high generic-closing rates (70\% and 61\%) but very different \textsc{Recovery} scores. \textsc{Recovery} is therefore not capturing a one-dimensional sycophancy axis along which all models can be ordered.

\begin{table}[ht]
  \caption{Per-model sycophancy proxy and \textsc{Recovery} weighted score used in the cross-correlation analysis. \emph{Generic} and \emph{Ungrnd.} are the two failure modes that operationalise sycophantic-helpfulness-without-substance; \emph{Syc.\ (within fail)} sums them as \% of \textsc{Recovery} failures.}
  \label{tab:sycophancy_proxy}
  \centering
  \small
  \smallskip
  \begin{tabular}{@{}lrrrrr@{}}
    \toprule
    \textbf{Model} & $n_\text{fail}$ & Generic \% & Ungrnd.\ \% & Syc.\ (within fail) \% & \textsc{Recovery} wt-score \% \\
    \midrule
    GPT-5.5             & 71  & 20 &  3 & 23 & 50.0 \\
    Qwen3.5-397B-A17B   & 116 & 61 &  3 & 64 & 26.6 \\
    Qwen3.5-9B          & 103 & 58 &  1 & 59 & 24.7 \\
    Qwen2.5-7B-Instruct & 145 & 12 &  5 & 17 & 15.4 \\
    GPT-4o              & 155 & 70 &  6 & 76 & 13.9 \\
    Qwen3-1.7B          & 150 & 11 &  5 & 16 & 12.8 \\
    Gemini-2.5-Pro      & 165 & 61 &  1 & 62 & 12.0 \\
    Llama-3.2-8B        & 174 & 51 &  6 & 57 & 5.0  \\
    \bottomrule
  \end{tabular}
\end{table}

\section{Long-context-recall analysis}
\label{app:longctx}

This appendix tests whether \textsc{Recovery} failures are primarily a retrieval-distance problem. If models were failing because the relevant detail was disclosed too many turns earlier, pass rates should fall as the distance between disclosure and trigger grows. We do not find that pattern for 15 of the 16 evaluated models, supporting the claim in Section~\ref{sec:per_stage} that long-context recall is not the main bottleneck.

\paragraph{Method.}
For every \textsc{Recovery} trigger and every evaluated model, we compute a \emph{disclosure distance}
\[
  d \;=\; t_\text{trigger} - t_\text{first-disclosure},
\]
where $t_\text{first-disclosure}$ is the earliest turn at which the rubric's pass criterion can be derived from the disclosed conversation. We estimate this turn by matching the rubric's pass criteria against earlier user messages and against the formal blueprint anchor list:
\begin{description}[leftmargin=10pt, itemsep=0pt, topsep=2pt]
  \item[Strong match.] Substantive token overlap $\geq 2$ between the rubric's pass criteria and an earlier user message or anchor. Used for $\sim$$87\%$ of triggers per model.
  \item[Weak match.] Substantive token overlap $\geq 1$. Used for $\sim$$7\%$ of triggers per model.
  \item[Fallback.] No match found; we attribute disclosure to $t-1$ (the user message immediately preceding the trigger). Used for the remaining $\sim$$6\%$ of triggers.
\end{description}
For each model we then regress the per-trigger weighted pass score (\textsc{Pass}=$1$, \textsc{Partial}=$0.5$, \textsc{Fail}=$0$) on $d$ via OLS and report the slope (the change in weighted pass rate per additional turn of disclosure distance) with a 1{,}000-resample trigger-level bootstrap CI. The aggregate sample is $\sim$$2{,}975$ \textsc{Recovery} triggers across the 16 models.

\paragraph{Results.}

Table~\ref{tab:longctx_slopes} reports the per-model slope and CI. For 15 of the 16 models, the confidence interval includes zero, so we do not see evidence that larger disclosure distance systematically lowers \textsc{Recovery} pass rate. The exception is Gemini-3.1-Pro: its slope is $-0.033$ per turn (95\% CI $[-0.066, -0.007]$), implying a roughly $3.3$ percentage-point decrease in weighted pass rate for each additional turn of disclosure distance. Across the realised range of $d$ (about 8 turns), this corresponds to a $\sim$$26$-pp drop. That effect is non-trivial, but it is still smaller than the $30$-pp+ frontier-vs-small gap on \textsc{Recovery}, and it is not the pattern seen across the model panel. 

\begin{table}[ht]
  \caption{Per-model disclosure-distance slope. \emph{Slope} is the change in weighted \textsc{Recovery} pass rate per additional turn of disclosure distance; values close to zero indicate that recall distance does not predict pass rate. 95\% CIs are 1{,}000-resample trigger-level bootstraps. Only Gemini-3.1-Pro has a CI strictly below zero, and its magnitude is smaller than the frontier-vs-small gap.}
  \label{tab:longctx_slopes}
  \centering
  \smallskip
  \begin{tabular}{@{}lcccc@{}}
    \toprule
    \textbf{Model} & $n$ & Mean $d$ & Slope (per turn) & 95\% CI \\
    \midrule
    GPT-5.5             & 191 & 5.85 & $-0.0037$ & $[-0.0388, +0.0312]$ \\
    Claude-Opus-4.7     & 191 & 5.85 & $-0.0008$ & $[-0.0372, +0.0341]$ \\
    Gemini-3.1-Pro      & 191 & 5.81 & $-0.0328$ & $[-0.0655, -0.0066]$ \\
    Gemini-2.5-Pro      & 191 & 5.85 & $-0.0018$ & $[-0.0278, +0.0211]$ \\
    Gemini-2.5-Flash    & 191 & 5.85 & $+0.0028$ & $[-0.0144, +0.0182]$ \\
    o4-mini             & 191 & 5.85 & $-0.0082$ & $[-0.0400, +0.0228]$ \\
    GPT-4o              & 191 & 5.85 & $-0.0155$ & $[-0.0443, +0.0119]$ \\
    Qwen3.5-397B-A17B   & 182 & 5.82 & $-0.0313$ & $[-0.0685, +0.0028]$ \\
    Kimi-K2.6           & 190 & 5.84 & $+0.0004$ & $[-0.0238, +0.0209]$ \\
    DeepSeek-V4-Flash   & 191 & 5.86 & $-0.0250$ & $[-0.0591, +0.0067]$ \\
    Llama-4-Maverick    & 191 & 5.85 & $-0.0024$ & $[-0.0164, +0.0094]$ \\
    MiMo-V2.5-Pro       & 191 & 5.85 & $-0.0118$ & $[-0.0434, +0.0167]$ \\
    Qwen3.5-9B          & 156 & 5.74 & $-0.0065$ & $[-0.0388, +0.0254]$ \\
    Qwen2.5-7B-Instruct & 191 & 5.85 & $-0.0128$ & $[-0.0396, +0.0151]$ \\
    Llama-3.2-8B        & 191 & 5.85 & $+0.0090$ & $[-0.0045, +0.0223]$ \\
    Qwen3-1.7B          & 191 & 5.85 & $-0.0005$ & $[-0.0211, +0.0200]$ \\
    \bottomrule
  \end{tabular}
\end{table}

\section{Logit Transform}
\label{app:logit}

This appendix spells out the logit transform used in Section~\ref{sec:logit}, explains the per-stage consistency check behind the shared-difficulty picture, and lists the per-model fit coefficients.

\paragraph{Score generation.}
For each model $m$ and configuration $c$, where $c$ is either one of the 24 communication styles or, in the per-stage analysis below, one of the three trigger types, let $k_{m,c}$ denote the number of triggers in cell $(m, c)$ scored \textsc{Pass}, and let $n_{m,c}$ denote the trigger count of that cell. The raw pass rate is $p_{m,c} = k_{m,c}/n_{m,c}$. We work in the Laplace-smoothed logit
\[
  \ell_{m,c} \;=\; \log\frac{k_{m,c} + 0.5}{n_{m,c} - k_{m,c} + 0.5},
\]
which keeps $\ell$ finite at the corner cases $k = 0$ and $k = n$ that several small models hit on \textsc{Recovery}. We define the shared difficulty axis for configuration $c$ as the cross-model mean $\bar\ell_c = M^{-1}\sum_m \ell_{m,c}$ over the $M = 16$ evaluated models, and fit each model's per-configuration logit against that axis by ordinary least squares: $\ell_{m,c} = a_m + b_m \bar\ell_c + \varepsilon_{m,c}$. The pass-rate inputs are the same as those reported in Table~\ref{tab:main_results} (Appendix~\ref{app:models}); for the per-style analysis we additionally aggregate within each of the 24 communication styles before applying the smoothed logit.

\begin{figure}[t]
  \centering
  \includegraphics[width=0.75\linewidth]{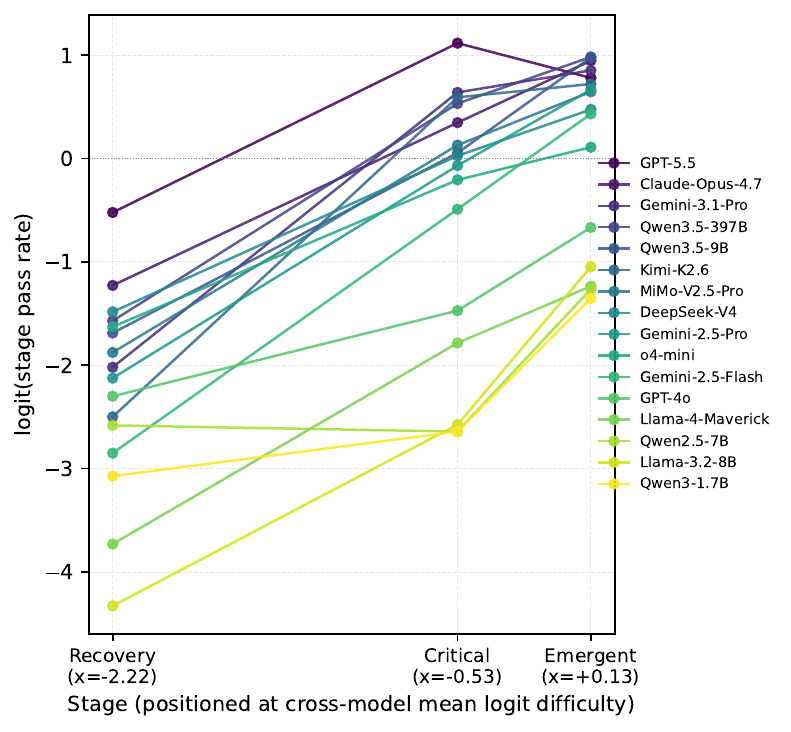}
  \caption{Per-model logit pass rates plotted against a shared stage-difficulty axis. Each polyline connects one model's \textsc{Emergent}, \textsc{Critical}, and \textsc{Recovery} logit pass rates; $x$-coordinates are the cross-model mean logit per stage (smaller $x$ = harder stage).}
  \label{fig:logit_stage}
\end{figure}

\paragraph{Per-stage logit consistency check.}
A complementary version of the same analysis aggregates at the \emph{stage} level rather than the style level. For each model $m$ and stage $s \in \{\textsc{Emergent}, \textsc{Critical}, \textsc{Recovery}\}$, we compute the Laplace-smoothed logit pass rate and place each stage on a shared $x$-axis at the cross-model mean logit ($x_{\textsc{E}}=+0.13$, $x_{\textsc{C}}=-0.53$, $x_{\textsc{R}}=-2.22$; smaller $x$ = harder stage). Connecting the three points per model produces 16 polylines (Figure~\ref{fig:logit_stage}). 

Under a strict shared-dimension hypothesis the polylines would be parallel: all models would show the same \textsc{Emergent}$\to$\textsc{Critical} gap and the same \textsc{Critical}$\to$\textsc{Recovery} gap, with between-model variation captured by an intercept for overall ability. This is a deliberately simplifying view, but it gives a useful baseline. Strong models tend to be strong across all stages (e.g.\ GPT-5.5), while weak models underperform throughout (e.g.\ Qwen3-1.7B).

More quantitatively, per-model $\textsc{Critical} \to \textsc{Recovery}$ slopes vary by a factor of ${\sim}2$ across the 16 models (slope mean $1.00$, std $0.45$, range $[-0.04, 1.83]$), while $\textsc{Emergent} \to \textsc{Critical}$ slopes are tighter (std $0.72$, range $[-0.51, 2.34]$). Concretely, the raw $\ell_{m,\textsc{C}} - \ell_{m,\textsc{R}}$ logit drop varies across models with std $0.76$. Top-\textsc{Recovery} models like GPT-5.5 stay relatively flat across the three stages, while several frontier models with strong \textsc{Critical} performance (Qwen3.5-397B-A17B, Kimi-K2.6, Gemini-3.1-Pro) collapse much more steeply on \textsc{Recovery}. 

\emph{This provides within-benchmark evidence that \textsc{Recovery} is not merely a harder rung on the same proactivity ladder. It complements the cross-benchmark decorrelation finding in Section~\ref{sec:benchselect}: the stage is difficult, but its difficulty is not uniform across models.}

\paragraph{Per-model fits.}
Table~\ref{tab:logit_fits} reports the per-model slope $b_m$ and $R^2$ on each of the four scoring axes. $b{=}1$ means the model lies exactly on the average style-difficulty line; $b{>}1$ means it amplifies style difficulty (better than average on easy styles, worse on hard ones); $b{<}1$ means it either absorbs style difficulty (uniform performance) or is at a floor (uniform failure).

\begin{table}[ht]
  \caption{Per-model linear fit of $\ell_{m,c}$ against the shared style-difficulty axis. $b$ is the slope, $R^2$ the coefficient of determination, computed across the 24 communication styles per axis.}
  \label{tab:logit_fits}
  \centering
  \setlength{\tabcolsep}{4pt}
  \smallskip
  \begin{tabular}{@{}lcc cc cc cc@{}}
    \toprule
    & \multicolumn{2}{c}{\textbf{Overall}} & \multicolumn{2}{c}{\textbf{Emergent}} & \multicolumn{2}{c}{\textbf{Critical}} & \multicolumn{2}{c}{\textbf{Recovery}} \\
    \cmidrule(lr){2-3} \cmidrule(lr){4-5} \cmidrule(lr){6-7} \cmidrule(lr){8-9}
    \textbf{Model} & $b$ & $R^2$ & $b$ & $R^2$ & $b$ & $R^2$ & $b$ & $R^2$ \\
    \midrule
    o4-mini            & 1.40 & 0.66 & 1.48 & 0.75 & 1.41 & 0.54 & 1.30 & 0.69 \\
    GPT-5.5            & 1.39 & 0.72 & 1.26 & 0.80 & 1.82 & 0.70 & 0.84 & 0.51 \\
    GPT-4o             & 1.39 & 0.55 & 1.06 & 0.55 & 1.25 & 0.45 & 0.96 & 0.65 \\
    Qwen3.5-397B       & 1.20 & 0.75 & 0.91 & 0.65 & 0.87 & 0.51 & 1.36 & 0.66 \\
    Qwen2.5-7B         & 1.16 & 0.20 & 1.14 & 0.51 & 0.50 & 0.10 & 0.72 & 0.26 \\
    Qwen3-1.7B         & 1.12 & 0.31 & 0.49 & 0.20 & 0.53 & 0.10 & 0.63 & 0.31 \\
    Kimi-K2.6          & 1.08 & 0.68 & 1.33 & 0.79 & 0.96 & 0.34 & 1.32 & 0.74 \\
    Qwen3.5-9B         & 0.90 & 0.38 & 1.07 & 0.70 & 1.30 & 0.43 & 1.10 & 0.58 \\
    Llama-4-Maverick   & 0.88 & 0.27 & 0.72 & 0.51 & 0.82 & 0.23 & 0.64 & 0.59 \\
    Gemini-2.5-Pro     & 0.84 & 0.42 & 1.00 & 0.56 & 1.12 & 0.53 & 1.04 & 0.61 \\
    Llama-3.2-8B       & 0.84 & 0.17 & 0.63 & 0.22 & 0.40 & 0.08 & 0.52 & 0.49 \\
    Gemini-3.1-Pro     & 0.81 & 0.56 & 1.19 & 0.63 & 0.90 & 0.54 & 1.08 & 0.58 \\
    MiMo-V2.5-Pro      & 0.81 & 0.44 & 0.95 & 0.66 & 0.89 & 0.40 & 1.34 & 0.74 \\
    DeepSeek-V4        & 0.78 & 0.51 & 0.95 & 0.87 & 1.16 & 0.55 & 0.97 & 0.49 \\
    Gemini-2.5-Flash   & 0.73 & 0.49 & 0.85 & 0.64 & 1.26 & 0.62 & 0.90 & 0.51 \\
    Claude-Opus-4.7    & 0.69 & 0.43 & 0.95 & 0.56 & 0.81 & 0.53 & 1.26 & 0.55 \\
    \bottomrule
  \end{tabular}
\end{table}

\section{Datasheet}
\label{app:datasheet}

This appendix documents how the released corpus is packaged and how reviewers can inspect it. The goal is to make the dataset usable without reverse-engineering the codebase: the release includes machine-readable Croissant metadata, explicit Responsible-AI fields, per-file schemas, and a conventional datasheet.

\paragraph{Croissant metadata.}
The released dataset ships with a machine-readable Croissant 1.0 description at \texttt{dataset/metadata.json}. This file is the canonical entry point for the corpus. It conforms to the MLCommons Croissant schema and includes the standard top-level fields (\texttt{name}, \texttt{description}, \texttt{license}, \texttt{url}, \texttt{version}, \texttt{datePublished}, \texttt{creator}, \texttt{citeAs}, \texttt{keywords}). The \texttt{distribution} block enumerates the six released JSONL files and the \texttt{recordSet} block declares one record set, \texttt{dialogues}, with the per-row schema typed against the Pydantic source of truth in \texttt{proactbench/types.py}.

\paragraph{Responsible AI fields (Croissant RAI extension).}
To make intended use, limitations, and provenance explicit, the metadata also includes the full set of MLCommons \emph{Responsible AI} fields under the \texttt{rai:} namespace:
\begin{description}[leftmargin=10pt, itemsep=0pt, topsep=2pt]
  \item[\texttt{rai:dataCollection}.] Synthetic; no human subjects or scraped user data. Personas are sampled from Nemotron-Personas-USA (CC-BY-4.0); scenarios, blueprints, and dialogues are generated by frontier LLMs (GPT-5.4 as Planner / User Agent, Gemini-2.5-Pro as curation assistant model).
  \item[\texttt{rai:dataAnnotationProtocol}.] Per-trigger Pass / Partial / Fail labels are produced by an LLM judge (GPT-5.4) acting as a neutral referee at turn $t{+}1$ against a rubric committed at turn $t$ (Section~\ref{sec:design}); labels include verbatim evidence quotes.
  \item[\texttt{rai:dataAnnotationAnalysis}.] Cross-family judge agreement reported in Appendix~\ref{app:judge_swap_appx} (Cohen's $\kappa$ 0.35--0.46 overall, all CIs strictly above zero); a human-validation study is described in Appendix~\ref{app:human_eval}.
  \item[\texttt{rai:dataPreprocessingProtocol}.] Independent-judge audit (Gemini-2.5-Pro) on every blueprint with four checks (blank-slate integrity, logical necessity, persona alignment, rubric clarity); only PASS blueprints proceed to dialogue curation.
  \item[\texttt{rai:dataReleaseMaintenancePlan}.] Hosted on Hugging Face; the GitHub repository carries the orchestration pipeline, generation prompts, and evaluation scripts. Maintenance commitment described in Appendix~\ref{app:licensing}.
  \item[\texttt{rai:dataUseCases}.] Intended uses: evaluating LLM proactivity (\textsc{Emergent} / \textsc{Critical} / \textsc{Recovery}); training-data substrate for proactivity preference learning; methodology study of multi-agent evaluation. Out-of-scope uses: deployment-readiness signal in non-US-English contexts, demographic-fairness benchmarking.
  \item[\texttt{rai:dataLimitation}.] US-English only; persona pool of 50; per-(model, style, type) cells are small ($\sim$8 dialogues per style on average); offline scoring re-uses the curation model's conversation history; the GPT-5.4-authored rubric is the single point of subjective interpretation.
  \item[\texttt{rai:dataBiases}.] Style filtering (24 of 64 binary CSI combinations) reflects authors' coherence judgement; persona category counts (Professional 46, Arts 45, Culinary 37, Sports 37, Travel 33) are imbalanced after audit drop-out (Appendix~\ref{app:personas}).
  \item[\texttt{rai:dataSocialImpact}.] Dual-use note: high \textsc{Recovery} scores should not be read as a universal training target; unsolicited initiative can be intrusive depending on user, privacy, and task context. We treat ProactBench as a capability probe.
  \item[\texttt{rai:personalSensitiveInformation}.] None. Personas are synthetic profiles from Nemotron-Personas-USA and carry no real-user PII; dialogues are generated by LLMs and contain no identifiable third-party data.
\end{description}
The metadata file validates against the official Croissant validator; reviewers can re-validate by running \texttt{mlcroissant validate dataset/metadata.json} after downloading the corpus.

\paragraph{Released files.}
The corpus is distributed as six JSONL files under \texttt{dataset/}. Each row is one typed record. The files separate task design, audit trail, and final dialogues, so users can inspect how each released conversation was produced. Per-file schemas live in \texttt{docs/DATA\_SCHEMAS.md} and the canonical Pydantic source in \texttt{proactbench/types.py}.
\begin{description}[leftmargin=10pt, itemsep=0pt, topsep=2pt]
  \item[\texttt{tasks.jsonl} (50 rows, 500 scenarios total).] One row per sampled persona; each row carries the persona's full set of Stage~1 candidate scenarios across the five life-domain categories (Section~\ref{sec:scenario}).
  \item[\texttt{selected\_tasks.jsonl} (19 rows, 25 scenarios total).] Curated scenario base selected for dialogue rollout: 19 personas with 25 (persona, category) scenarios stratified at five per life-domain category.
  \item[\texttt{blueprints.jsonl} (250 rows).] All Stage 2 blueprints (turn-by-turn interaction plans), one per (selected scenario $\times$ communication-style) pairing: $25 \times 10 = 250$.
  \item[\texttt{validation\_results.jsonl} (250 rows).] Independent-judge audit decisions (PASS / NEEDS\_REFINEMENT / FAIL) with the four sub-checks for every blueprint.
  \item[\texttt{validated\_blueprints.jsonl} (207 rows).] Audit-passing subset, the input to dialogue curation.
  \item[\texttt{final\_dialogues.jsonl} (198 rows).] The released benchmark corpus: 198 dialogues, 624 trigger points (201 \textsc{Emergent} / 232 \textsc{Critical} / 191 \textsc{Recovery}), each with the full ten-turn record, the trigger schedule, and the Planner-authored rubrics.
\end{description}

\paragraph{Datasheet.}
A human-readable datasheet following the \emph{Datasheets for Datasets} structure (Gebru et al., 2021) ships at \texttt{dataset/DATASHEET.md}. It covers eight sections: Motivation, Composition, Collection process, Preprocessing, Uses, Distribution, Maintenance, and Citation. The datasheet mirrors the Croissant RAI material above in prose for reviewers who prefer a narrative account of assumptions, intended uses, and limitations.

\section{Licensing and CITATION}
\label{app:licensing}

This appendix explains the release terms, citation path, and responsible-use guidance for downstream users.

\paragraph{Licence split.}
The release has two licence layers. The orchestration pipeline (\texttt{proactbench/}), generation prompts, evaluation scripts, and released JSONL corpus are distributed under the \textbf{Apache 2.0} licence. Persona-derived content (the \texttt{uuid} identifiers in \texttt{tasks.jsonl} and the multi-aspect persona text rendered at runtime by \texttt{build\_global\_persona}) inherits the upstream \textbf{Nemotron-Personas-USA CC-BY-4.0} licence \citep{meyer2025nemotronpersonas}. 

Downstream users who re-render the multi-aspect persona for their own evaluations must preserve NVIDIA's attribution per CC-BY-4.0; the released corpus itself only redistributes \texttt{uuid}s, never the raw persona text. The two licences are mutually compatible; Apache 2.0 imposes a permissive grant on the code and corpus while CC-BY-4.0 governs the persona text only, and nothing in Apache 2.0 prevents a CC-BY-4.0 attribution requirement from being honoured downstream.

\paragraph{Hosting and access.}
Code and data are mirrored at \url{https://anonymous.4open.science/r/ProactBench-81A3/README.md} during double-blind review (read-only, anonymous). The non-anonymous \textbf{GitHub} repository and the \textbf{Hugging Face} dataset URL will be disclosed in the camera-ready version. The Croissant metadata file (\texttt{dataset/metadata.json}, Appendix~\ref{app:datasheet}) is the canonical entry point; every released file under \texttt{dataset/} is enumerated there with a stable URL.

\paragraph{CITATION.cff.}
A \texttt{CITATION.cff} file at the GitHub root exposes the canonical citation for the dataset and the code in the format consumed by GitHub's ``Cite this repository'' button and by tools such as Zenodo and Zotero. Bibtex export is available via the \texttt{citeAs} field of \texttt{dataset/metadata.json}.


\paragraph{Ethics statement.}
ProactBench evaluates LLM proactivity in a synthetic conversational setting; no human subjects are involved in the evaluation pipeline, no real-user data is collected or redistributed, and the personas are synthetic with no PII (Appendix~\ref{app:personas}). The human-validation study (Appendix~\ref{app:human_eval}) is conducted under an IRB-approved protocol with informed consent, Prolific-compensated annotators, and no collection of personal information beyond platform IDs. The dual-use consideration (high proactivity is not a universal training target; unsolicited initiative can be intrusive) is also flagged in \texttt{rai:dataSocialImpact} of the Croissant metadata and in the Discussion.

\paragraph{Responsible-use guidance.}
We recommend that downstream users of ProactBench:
\begin{description}[leftmargin=10pt, itemsep=0pt, topsep=2pt]
  \item[Treat \textsc{Recovery} as a capability probe, not a deployment target.] Optimising directly for \textsc{Recovery} pass rate without paired sycophancy / over-helpfulness controls risks producing models that volunteer initiative when users would prefer reticence.
  \item[Do not use ProactBench scores as a deployment-readiness signal beyond US-English contexts.] Norms around unsolicited advice, directness, and initiative-taking differ substantially across cultures (Discussion); the persona pool is US-only, and the communication-style instrument (CSI) is most extensively validated in workplace English.
  \item[Cite the upstream Nemotron-Personas-USA dataset.] Per its CC-BY-4.0 terms, any work building on the persona text must attribute NVIDIA.
\end{description}

\end{document}